\documentclass[acmtog]{acmart}

\usepackage{multirow}
\usepackage{makecell}
\usepackage{float}
\usepackage{xcolor,colortbl}
\usepackage{bbm}
\usepackage{algorithm}
\usepackage{algpseudocode}

\AtBeginDocument{%
  }

\definecolor{firstcolor}{rgb}{0.95,0.69,0.51}
\definecolor{secondcolor}{rgb}{0.99,0.9,0.6}
\definecolor{thirdcolor}{rgb}{0.99,0.90,0.60}
\newcommand{\cf}[1]{\cellcolor{firstcolor} #1}
\newcommand{\cs}[1]{\cellcolor{secondcolor} #1}
\newcommand{\uli}[1]{\underline{#1}}

\begin{document}

\title{RRG-SLAM: Real-time Reflection-aware Gaussian SLAM for Indoor
Scenes}


\author{Yong Liu}
\affiliation{%
  \institution{State Key Laboratory of CAD\&CG, Zhejiang University}
  \country{China}
}
\email{yongliu6@zju.edu.cn}
\orcid{0009-0000-8133-5686}

\author{Keyang Ye}
\affiliation{%
  \institution{State Key Laboratory of CAD\&CG, Zhejiang University}
  \country{China}
}
\email{yekeyang@zju.edu.cn}
\orcid{0009-0005-8675-566X}

\author{Zhexi Peng}
\affiliation{%
  \institution{State Key Laboratory of CAD\&CG, Zhejiang University}
  \country{China}
}
\email{zhexipeng@zju.edu.cn}
\orcid{0000-0003-4342-5263}

\author{Ruixian Mei}
\affiliation{%
  \institution{State Key Laboratory of CAD\&CG, Zhejiang University}
  \country{China}
}
\email{mrx@zju.edu.cn}
\orcid{0009-0009-2757-1376}

\author{Kun Zhou}
\affiliation{%
  \institution{State Key Laboratory of CAD\&CG, Zhejiang University}
  \country{China}
}
\email{kunzhou@acm.org}
\orcid{0000-0003-4243-6112}

\author{Tianjia Shao}
\affiliation{%
  \institution{State Key Laboratory of CAD\&CG, Zhejiang University}
  \country{China}
}
\email{tjshao@zju.edu.cn}
\orcid{0000-0001-5485-3752}

\begin{abstract}
We introduce the first real-time reflection-aware Gaussian SLAM system for indoor scenes. The system features a reflection-aware TSDF-Gaussian hybrid representation that explicitly separates diffuse scene appearance from reflection components. The base scene is modeled by a TSDF volume and a set of base Gaussians capturing geometry and diffuse appearance, while planar reflections are represented by reflection Gaussian groups associated with detected reflective planes.
The rendering is performed in three passes: TSDF raycasting first yields surface color, depth, plane IDs and reflection masks; base Gaussians are then rendered order-independently with depth culling and combined with the TSDF output to form the base image; finally, under the guidance of the plane ID map, reflection Gaussians from different reflection groups are rasterized only into their corresponding planar regions to generate the reflection image, which is subsequently composited with the base image via the reflection mask to produce the final output.
For online reconstruction, our system first estimates the camera pose through reflection-aware tracking to suppress interference of reflection-dominated regions. It then identifies reflective planes using geometric, semantic, and temporal cues, and fuses the observations into the augmented TSDF volume with reflection-aware attributes. Afterwards the base and reflection Gaussians are initialized, optimized, and pruned online to maintain both reconstruction quality and efficiency. Experiments on a variety of datasets show that our method outperforms existing SLAM systems in reconstruction quality, tracking robustness, and novel-view rendering for indoor environments with reflections, while preserving real-time performance.
\end{abstract}

\ccsdesc[500]{Computing methodologies~Point-based models}
\ccsdesc[500]{Computing methodologies~Reconstruction}

\keywords{RGB-D SLAM, 3D Gaussian Splatting, Real-Time Reconstruction}
\begin{teaserfigure}
  \includegraphics[width=\textwidth]{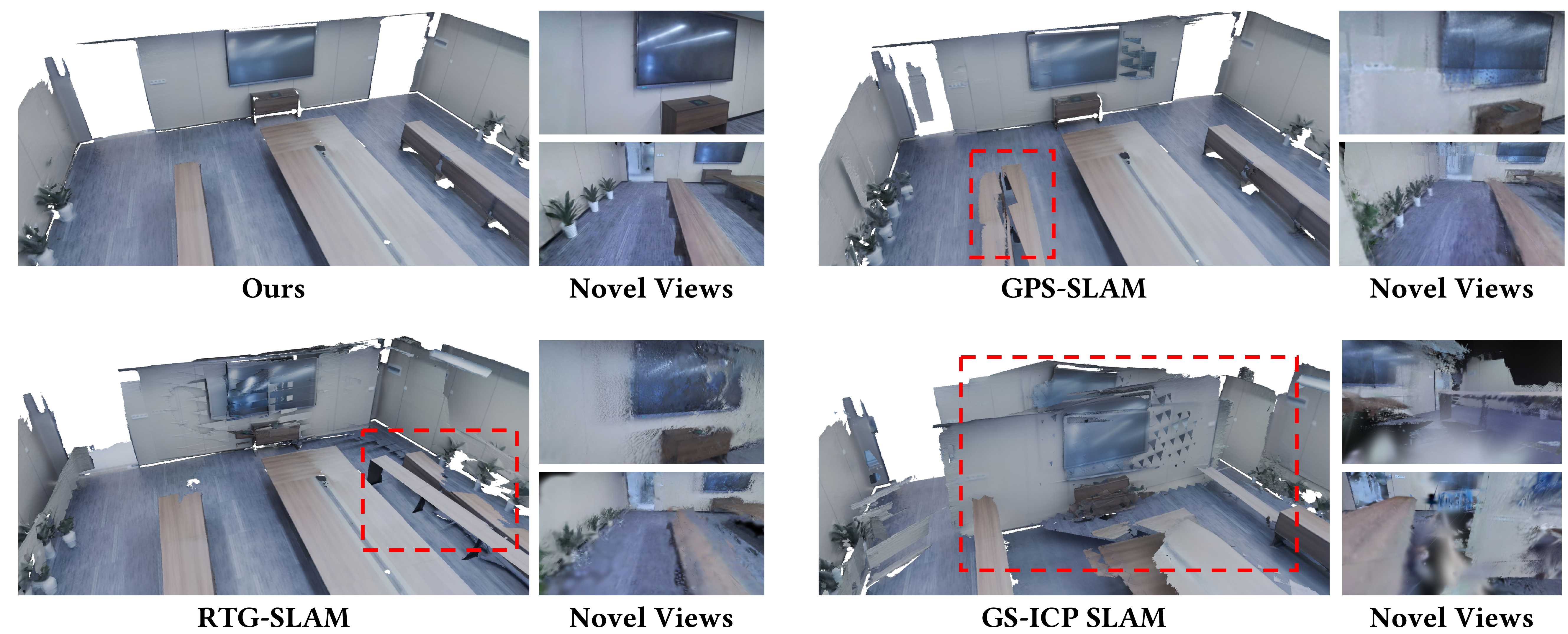}
  \caption{A meeting room containing strong planar reflections from a screen and polished tables. For clearer visualization, we crop out the ceiling and glass doors. We compare our method with state-of-the-art Gaussian SLAM approaches (GPS-SLAM~\cite{gps-slam}, RTG-SLAM~\cite{rtg-slam}, and GS-ICP SLAM~\cite{gs-icp}) for mesh reconstruction and novel-view synthesis. Regions with large drifting errors are highlighted with red boxes. Our method enables more stable tracking and improves rendering quality in reflective environments.}
  \Description{Comparison of Gaussian SLAM methods in a reflective indoor scene with reconstructed geometry and rendered novel views.}
  \label{fig:teaser}
\end{teaserfigure}


\maketitle

\section{INTRODUCTION}
Real-time 3D indoor scene reconstruction is a fundamental problem in computer vision and robotics, with broad applications in robotics and augmented reality. 
Unlike outdoor settings, indoor environments often contain complex appearance effects caused by artificial lighting and planar reflective surfaces, such as polished wooden floors, tiled floors, and display screens.
Reconstructing such indoor environments with strong planar reflections in real time is particularly challenging. 
Classical RGB-D SLAM systems~\cite{kinectfusion,bundlefusion,orbslam} mainly focus on geometry reconstruction, without modeling the photorealistic appearance.  
Recent 3D-Gaussian-based RGB-D SLAMs~\cite{3dgs, splatam, gs-icp,rtg-slam,gps-slam} have enabled real-time 3D indoor reconstruction with high-fidelity appearance. 
However, these methods all assume that indoor environments consist of purely diffuse materials. When reflection phenomena exist in indoor scenes, their 3D Gaussian representations struggle to accurately capture such reflected content, leading to apparent artifacts in the reconstructed reflective regions.
More critically, the strong reflections can violate photometric consistency and corrupt image features, making photometric- and feature-based tracking unreliable. As a result, existing SLAM methods often exhibit large drifting errors in indoor scenes with strong reflections (see Fig.~\ref{fig:teaser} for example).

In this paper, we introduce a reflection-aware scene representation for planar reflections, which explicitly separates base scene appearance from reflection components, while
remaining compatible with online SLAM.
The representation consists of a truncated signed distance field (TSDF) volume and a set of base 3D Gaussians capturing geometry and diffuse appearance, as well as a set of reflection groups modeling plane-induced reflective effects.
Specifically, the TSDF volume models the scene geometry and coarse diffuse appearance, while base Gaussians refine the diffuse appearance beyond the TSDF volume. A reflection group comprises a reflective plane and its associated reflection Gaussians. This design is motivated by the symmetry of planar reflections: reflected content can be interpreted as virtual scene behind the corresponding plane and represented by Gaussians in this virtual space.
The TSDF is further augmented with reflection-related attributes: plane association recording the detected plane each voxel belongs to, reflection strength marking reflective regions, and temporal color variance capturing color changes accumulated over repeated observations. These attributes support reflection rendering and reflective plane detection. 

The rendering process follows a three-pass design. The first pass raycasts the TSDF volume to obtain the TSDF color, surface depth, plane ID map, and reflection mask. The plane ID map assigns each valid pixel to a detected planar region, while the reflection mask labels reflective pixels for compositing. The second pass renders base Gaussians with order-independent accumulation and depth culling using the raycast surface depth, and combines their accumulated color with the TSDF color to produce the base color image. The third pass uses our customized rasterizer to render the reflection Gaussians from all reflection groups in a single pass. Guided by the raycast plane ID map, each group contributes only to the pixels assigned to its associated plane, producing the reflection color image without interference across groups. The final image is obtained by compositing the base and reflection colors using the reflection mask.

To make this representation practical for online reconstruction, we need to reliably identify reflective planes. A single cue is insufficient: depth reveals planar structures but cannot determine whether they are reflective, while image semantics suggests potentially reflective surfaces but does not verify reflection effects. We therefore combine geometry and semantics with a temporal cue from reconstruction. Specifically, we use CAPE~\cite{cape} to segment planar regions from the input depth map, and Grounded SAM2~\cite{groundingdino,sam2} to detect predefined categories that are likely to exhibit reflections in the input color image. Their association gives candidate reflective planes, which are further verified using a raycast color variance map whose voxel-wise statistics are updated during TSDF fusion. Planes with sufficiently large high-variance regions are marked as reflective in the current detection window. To avoid noisy single-frame detections, we aggregate recent decisions in a sliding window and accept a plane as reflective only when it is repeatedly identified across multiple frames.

Our reflection-aware representation also benefits tracking by explicitly excluding reflection-dominated regions. We first estimate an initial pose using point-to-plane ICP, which aligns the current depth observation to the raycast TSDF surface and provides a robust geometric initialization. Given this pose, we render the reflection color image from our representation and threshold its intensity to obtain a binary tracking mask that identifies reflection-dominated pixels. In the second stage, we perform feature-based pose refinement on the unmasked regions, excluding reflection-dominated pixels from feature extraction. This allows the tracker to benefit from reliable image features while avoiding reflection-induced interference.

Building on the proposed reflection-aware representation and its associated online operations, we construct the first RGB-D Gaussian SLAM system designed for indoor scenes with planar reflections. Given an input RGB-D frame, our system performs reflection-aware tracking to estimate the camera pose, identifies reflective planes using geometric, semantic, and temporal cues, and fuses the observations into the TSDF volume. The scene is further refined through Gaussian reconstruction, where base and reflection Gaussians are initialized with carefully designed strategies, optimized online, and pruned to keep the representation compact. We evaluate our method on self-captured reflective scenes and a synthetic reflective dataset, and compare it with state-of-the-art Gaussian-based SLAM methods. Our system achieves better reconstruction quality, more robust tracking, and higher-quality novel-view rendering in scenes with reflections while maintaining real-time performance. We also evaluate on public RGB-D benchmarks, including ScanNet++, Replica, and TUM-RGBD, to demonstrate its advantages in scenes with reflections as
well as its compatibility with general indoor scenes.
In summary, our contributions are:
\begin{itemize}
\item We propose the first real-time RGB-D Gaussian SLAM system designed for indoor scenes with planar reflections.
\item We propose a novel hybrid representation combining TSDF, base Gaussians, and reflection Gaussians, integrated with real-time reflection detection, reflection-aware tracking, and online optimization to enable robust online reconstruction of scenes with reflections.
\item We demonstrate through extensive experiments that our system achieves state-of-the-art reconstruction quality and robust tracking accuracy on both public and self-captured datasets while maintaining real-time performance.
\end{itemize}

\section{RELATED WORK}

\paragraph{Classical RGB-D SLAM}

Classical RGB-D SLAM systems mainly differ in the tracking strategies they employ, which typically rely on geometric alignment, photometric consistency, or sparse image features. For instance, KinectFusion-style methods~\cite{kinectfusion,voxelhashing,steinbrucker2013large,kahler2015very} estimate camera poses via frame-to-model ICP, while systems such as ElasticFusion~\cite{elasticfusion} and BundleFusion~\cite{bundlefusion} incorporate both geometric and photometric terms in a joint optimization framework. Feature-based approaches~\cite{endres2012evaluation,labbe2019rtab,sumikura2019openvslam}, such as ORB-SLAM~\cite{orbslam,orbslam2,orbslam3}, instead track sparse keypoints and refine poses through bundle adjustment.
While these methods achieve strong performance in general environments, they do not explicitly account for reflection-induced appearance variations. In scenes with prominent planar reflections, view-dependent effects can degrade the reliability of photometric alignment and feature matching, leading to unstable pose estimation. In contrast, our method introduces a reflection-aware tracking strategy that explicitly identifies and suppresses reflection-dominated regions during refinement, improving robustness in such challenging scenarios.

\paragraph{Radiance Field-based RGB-D SLAM} 

Radiance-field-based RGB-D SLAM extends scene reconstruction beyond purely geometric mapping toward photorealistic scene representation and novel-view synthesis. Early NeRF-based SLAM systems, such as iMAP~\cite{imap}, NICE-SLAM~\cite{nice-slam}, ESLAM~\cite{eslam}, Co-SLAM~\cite{co-slam}, and Point-SLAM~\cite{point-slam}, demonstrated that jointly optimizing camera poses and a neural radiance field can significantly improve rendering quality compared with classical RGB-D SLAM systems. However, these methods rely on repeated MLP evaluation and volumetric rendering, which are computationally expensive and therefore limit their efficiency for online operation.
More recently, 3D Gaussian Splatting (3DGS)~\cite{3dgs} introduced an explicit radiance representation based on Gaussian primitives together with differentiable rasterization, enabling much faster rendering while maintaining high visual fidelity. Motivated by these advantages, a growing number of Gaussian-based SLAM systems~\cite{gs-slam,cg-slam,yugay2023gaussian,gaus-slam,rtg-slam,gs-icp,splatam,gps-slam} have adopted 3DGS as the scene representation. Compared with NeRF-based approaches, these methods are substantially more efficient and better suited for online reconstruction, while also achieving higher-quality novel-view synthesis than classical RGB-D SLAM.
Among them, GPS-SLAM~\cite{gps-slam} is particularly relevant to our work. Instead of relying solely on Gaussians, it combines a TSDF volume with sparse Gaussians in a hybrid representation, where the TSDF provides coarse geometry and appearance, and the Gaussians capture finer details. By adopting a sorting-free two-pass rendering strategy, GPS-SLAM achieves very high frame rates without sacrificing rendering quality. This design makes it especially attractive for real-time RGB-D SLAM, and also provides a strong foundation for further extending the representation toward more challenging scene effects.

Despite these advances, existing radiance-field-based RGB-D SLAM methods generally assume that scene appearance can be represented by standard radiance primitives, such as MLP-based fields or SH-parameterized Gaussians. In strongly reflective scenes, however, this assumption becomes insufficient. Reflections introduce complex view-dependent appearance that is difficult to model accurately, which in turn affects both pose tracking and scene mapping. To address this limitation, we extend the hybrid TSDF-Gaussian representation with explicit reflection modeling. Our representation consists of a TSDF volume for coarse geometry and color, a set of base Gaussians for refining the stable scene appearance, and a set of plane-associated reflection Gaussians for modeling reflective content induced by planar surfaces. This design makes the representation better suited to indoor reflective scenes, improving both tracking robustness and mapping quality, while also benefiting novel-view synthesis.

\paragraph{Offline Reflection Reconstruction} 


Reflection reconstruction has also been widely studied in offline settings. Existing methods can be broadly categorized into environment-map-based, ray-tracing-based, and geometry-aware approaches. Environment-map-based methods~\cite{boss2021nerd,3dgs-dr,jiang2024gaussianshader,zhang2022differentiable,zhang2021nerfactor}, such as 3DGS-DR~\cite{3dgs-dr} and GaussianShader~\cite{jiang2024gaussianshader}, model reflections through environment lookups, but their distant-light assumption is often insufficient for indoor scenes with strong near-field effects. Ray-tracing-based methods~\cite{gao2024planar,verbin2024nerf,envgs} model reflection more faithfully through secondary ray sampling, but their computational cost is too high for real-time operation. Geometry-aware methods, such as Mirror-NeRF~\cite{zeng2023mirror}, Mirrorgaussian~\cite{liu2024mirrorgaussian}, and TR-Gaussian~\cite{liu2026tr}, exploit planar symmetry to reconstruct virtual content behind reflective surfaces and achieve high-quality results. Among these, geometry-aware methods are most relevant to our work because they explicitly exploit planar structure for reflection modeling. However, these methods are designed for offline reconstruction rather than online SLAM. They typically rely on dense multi-view observations and iterative optimization over the full image set, making them difficult to apply directly in a real-time setting. To bridge this gap, we design an online reflection-aware reconstruction pipeline for RGB-D SLAM. In particular, we introduce online reflective plane detection together with a tailored reflection Gaussian initialization strategy, which significantly reduces the difficulty of reflection modeling and enables explicit reconstruction from sequential input while preserving real-time performance.

\section{Reflection-Aware TSDF-Gaussian Hybrid Representation}

\begin{figure*}[ht]
    \centering
    \includegraphics[width= \textwidth]{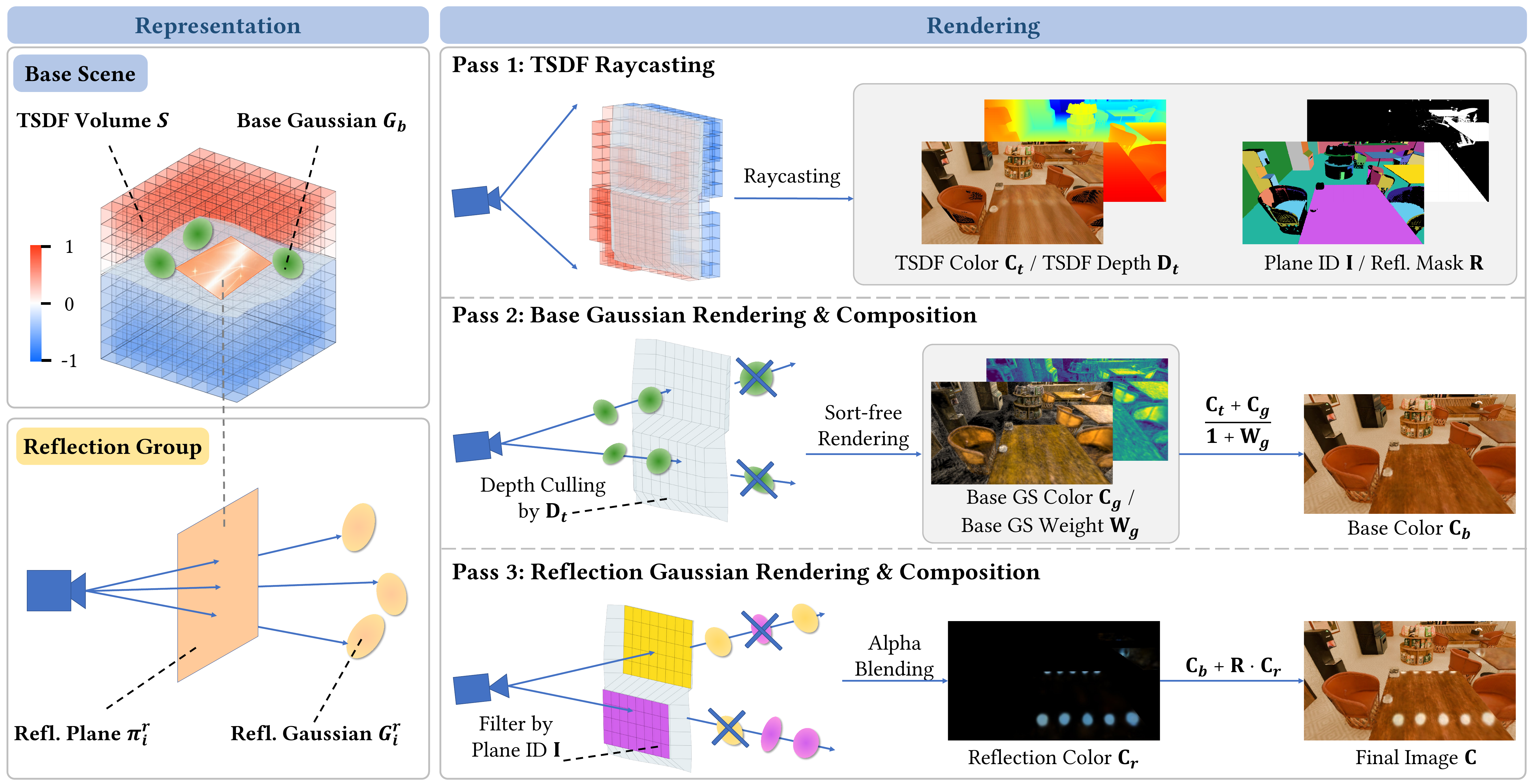}
    \Description{
    A schematic overview of the proposed reflection-aware scene representation and rendering pipeline. The figure shows a base scene represented by a TSDF volume and base Gaussians, together with plane-associated reflection groups. The rendering pipeline consists of TSDF raycasting, base Gaussian rendering, and reflection Gaussian rendering, which are composited to produce the final image.
    }
    \caption{Overview of our reflection-aware scene representation and three-pass rendering pipeline. The representation decomposes the scene into a base scene (TSDF volume and base Gaussians) capturing geometry and diffuse appearance, and reflection groups modeling planar reflections. Rendering proceeds in three passes: (1) TSDF raycasting produces color, depth, plane ID, and reflection mask; (2) base Gaussians are rendered with depth culling and combined with TSDF color to produce the base color; (3) reflection Gaussians are rendered conditioned on their associated plane IDs and composited using the reflection mask to generate the final image. This decomposition enables explicit modeling of reflective content while maintaining efficient rendering of the base scene.}
    \label{fig:rendering}
\end{figure*}

\subsection{Modeling}

We propose a reflection-aware scene representation that explicitly separates base scene appearance from reflection components, while remaining compatible with online SLAM.
Our representation consists of two parts: a base scene representation that captures geometry and diffuse appearance, and a set of reflection groups that model plane-induced reflective effects. This design is particularly suitable for indoor environments, where reflections are often associated with dominant planar surfaces such as tables, screens, and floors.

\paragraph{Base Representation.}
Our base representation builds upon GPS-SLAM~\cite{gps-slam}. Following its hybrid design, we use a TSDF volume $\mathcal{S}$ to represent scene geometry and coarse diffuse appearance and a set of base Gaussians $\mathcal{G}_b$ to refine diffuse appearance. Each TSDF voxel $\mathbf{p}$ stores its signed distance $d(\mathbf{p})$ and color $\mathbf{c}(\mathbf{p})$. We further augment each voxel with three reflection-aware attributes: a color variance accumulator $S_c(\mathbf{p})$ for capturing temporal appearance variation, a reflection strength $r(\mathbf{p})$ marking the reflective regions, and a plane ID $\pi(\mathbf{p})$ that associates the voxel with a planar surface.
The base Gaussians are defined as $\mathcal{G}_b=\{\mathbf{p}_{i}^{b}, \mathbf{s}_{i}^{b}, \mathbf{r}_{i}^{b}, \sigma_{i}^{b}, \mathbf{SH}_{i}^{b}\}_{i=1}^{N_b}$, parameterized by position, scale, rotation, opacity, and SH coefficients.

\paragraph{Reflection Groups.}
To explicitly model planar reflections, we organize reflective appearance into plane-level reflection groups. During scanning, planar surfaces are detected online and represented as $\pi_i=(\mathbf{n}_i,b_i)$ with a unique ID $i$. The plane ID is further propagated to TSDF voxels during fusion, allowing the reconstructed geometry to maintain consistent plane associations.
Each detected reflective plane defines a reflection group $\{\pi_i^r,\mathcal{G}_i^r\}$, which consists of the reflective plane $\pi_i^r$ and its associated reflection Gaussian set $\mathcal{G}_{i}^r$. The complete reflection Gaussian set is denoted by: 
\begin{equation}
   \mathcal{G}_r = \bigcup_i \mathcal{G}_i^r
\end{equation}

This design is motivated by the symmetry of planar reflection. Specifically, reflected appearance can be interpreted as virtual content behind the reflective plane and represented using Gaussians in this virtual space. Reflection Gaussians share the same parameterization as base Gaussians, while each Gaussian additionally stores the ID of its associated reflective plane:
\begin{equation}
\mathcal{G}_{i}^r=\{\mathbf{p}_j^r,\mathbf{s}_j^r,\mathbf{r}_j^r,\sigma_j^r,\mathbf{SH}_j^r,\ell_j^r\}_{j=1}^{N_{i}^r},
\end{equation}
where $\ell_j^r$ denotes the reflective plane associated with the Gaussian. We also maintain a reflective plane lookup table $L_{ref}$ to record whether each detected plane has been identified as reflective.

Organizing reflections at the plane level provides two advantages. First, it disentangles reflective appearance from the base scene representation, preventing reflection-induced artifacts from contaminating diffuse reconstruction. Second, it enables efficient plane-conditioned rendering, where each reflection group contributes only to pixels associated with its corresponding plane.

\subsection{Rendering}

\begin{figure*}[t]
    \centering
    \includegraphics[width= \textwidth]{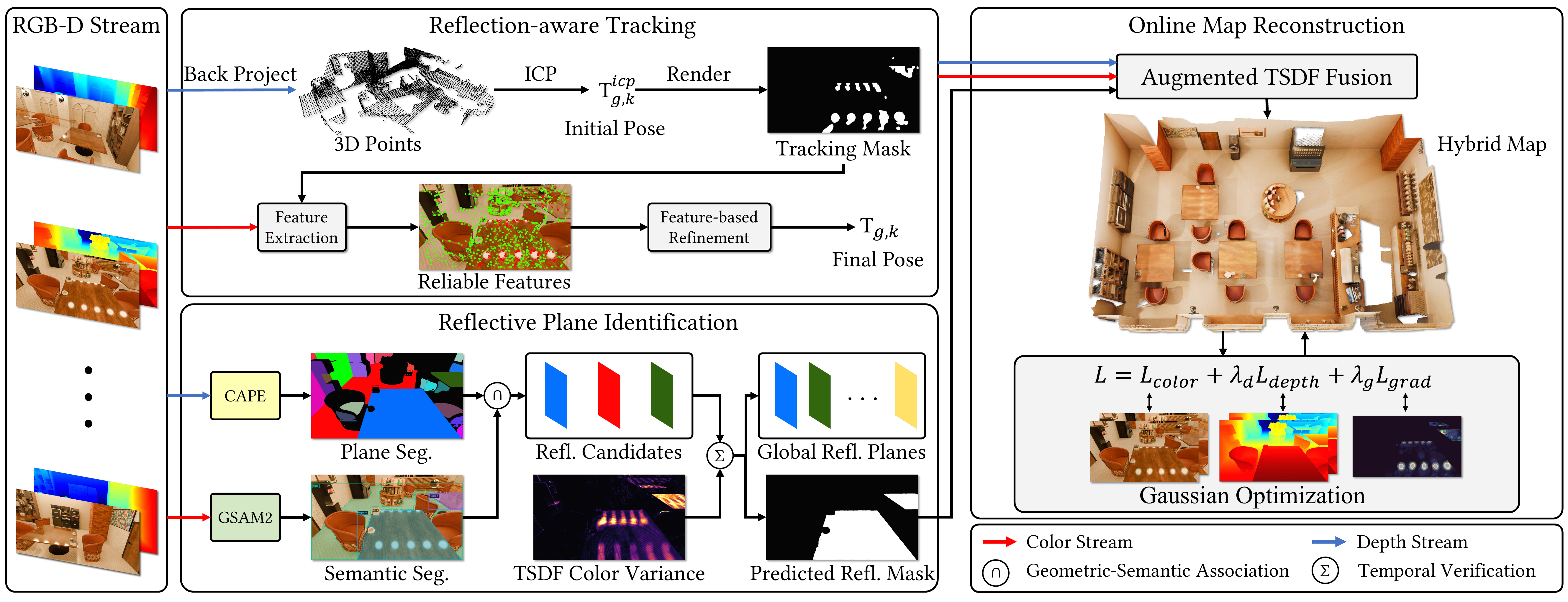}
    \Description{
    A pipeline diagram of a reflection-aware RGB-D SLAM system. The left side shows an RGB-D input stream. The top middle block illustrates reflection-aware tracking, including depth back-projection, ICP-based initial pose estimation, reflection-mask rendering, reliable feature extraction, and feature-based pose refinement. The bottom middle block shows reflective plane identification, where CAPE plane segmentation, GSAM2 semantic segmentation, and TSDF color variance are combined through geometric-semantic association and temporal verification to obtain global reflective planes and a predicted reflection mask. The right block shows hybrid map reconstruction, where color, depth, and reflection-mask information are fused into an augmented TSDF volume and used for Gaussian optimization.
    }
    \caption{Overview of the proposed reflection-aware RGB-D SLAM pipeline. Given an RGB-D stream, the system performs reflection-aware tracking, reflective plane identification, and online map reconstruction. The tracking module first estimates an initial pose using ICP on depth-derived 3D points, and then refines the pose using reliable color features selected by the rendered tracking mask. The reflective plane identification module combines geometric plane segmentation, semantic segmentation, and TSDF color variance to detect global reflective planes and generate a predicted reflection mask. Finally, the online map reconstruction performs augmented TSDF fusion and Gaussian optimization.}
    \label{fig:pipeline}
\end{figure*}

As illustrated in Fig.~\ref{fig:rendering}, our rendering pipeline consists of three passes: TSDF raycasting, base Gaussian rendering, and reflection Gaussian rendering.

In the first pass, we raycast the TSDF volume to obtain a TSDF color image $\mathbf{C}_t$ and a depth map $\mathbf{D}_t$. In addition, we extract a plane ID map $\mathbf{I}$ and a reflection strength map $\mathbf{R}_{raw}$. The plane ID map assigns each visible pixel to its associated planar surface, while $\mathbf{R}_{raw}$ measures the likelihood of reflective effects at each pixel. We further binarize $\mathbf{R}_{raw}$ using a threshold of $0.5$ to obtain the reflection mask $\mathbf{R}$.

In the second pass, we render the base Gaussians $\mathcal{G}_b$. Since base Gaussians correspond to directly observed scene surfaces, we perform depth culling against the raycast TSDF surface depth $\mathbf{D}_t$. The Gaussian attributes are accumulated in an order-independent manner similar to~\cite{ges}:
\begin{equation}
\mathcal{A}_b[\mathbf{x}](\mathbf{u})
 =
\sum_{i=1}^{K}
\mathbbm{1}\left(d_i^b < \mathbf{D}_t(\mathbf{u})+\epsilon\right)
\alpha_i^b(\mathbf{u})\mathbf{x}_i^b,
\label{eq:operator}
\end{equation}
where
\begin{equation}
\alpha_i^b(\mathbf{u})=
\sigma_i^b
\exp\left(
-\frac{
(\mathbf{u}-\hat{\mathbf{p}}_i^b)^T
(\boldsymbol{\Sigma}_{i}^b)^{-1}
(\mathbf{u}-\hat{\mathbf{p}}_i^b)
}{2}
\right).
\end{equation}

Here $\mathbf{u}$ denotes the pixel coordinate, $\mathbbm{1}(\cdot)$ is the indicator function, and $\epsilon$ is the truncation distance~\cite{gps-slam}. For the $i$-th Gaussian, $d_i^b$ and $\alpha_i^b$ denote its projected depth and weight, respectively. $\boldsymbol{\Sigma}_{i}^b$ is the covariance matrix of the projected Gaussian in image space, and $\hat{\mathbf{p}}_i^b$ is the projected Gaussian center. The term $\mathbf{x}_i^b$ denotes an arbitrary Gaussian attribute. By substituting $\mathbf{x}_i^b$ with the view-dependent color $\mathbf{c}_i^b$, scalar unity $1$, or Gaussian depth $d_i^b$, we obtain the accumulated color $\mathbf{C}_g=\mathcal{A}_b[\mathbf{c}]$, accumulated weight $\mathbf{W}_g=\mathcal{A}_b[1]$, and Gaussian depth map $\mathbf{D}_g=\mathcal{A}_b[d_i]$, respectively. The final base color image and depth map are computed as
\begin{equation}
\mathbf{C}_b=
\frac{\mathbf{C}_t+\mathbf{C}_g}
{1+\mathbf{W}_g},
\qquad
\mathbf{D}_b=
\frac{\mathbf{D}_t+\mathbf{D}_g}
{1+\mathbf{W}_g}.
\label{eq:base_render}
\end{equation}

In the third pass, we render all reflection Gaussians in a single pass to obtain the reflection image $\mathbf{C}_r$. To ensure that each reflection group contributes only to its associated reflective plane, we implement plane-conditioned rendering as a per-pixel filtering operation during rasterization:
\begin{equation}
\mathcal{A}_r[\mathbf{x}](\mathbf{u})=
\sum_{i=1}^{K}
\prod_{j=1}^{i-1}
(1-\alpha_j^r(\mathbf{u}))
\alpha_i^r(\mathbf{u})
\mathbf{x}_i^r,
\label{eq:reflection_operator}
\end{equation}
where
\begin{equation}
\alpha_i^r(\mathbf{u})=
\mathbbm{1}\left(
\ell_i^r=\mathbf{I}(\mathbf{u})
\right)
\sigma_i^r
\exp\left(
-\frac{
(\mathbf{u}-\hat{\mathbf{p}}_i^r)^T
(\boldsymbol{\Sigma}_{i}^{r})^{-1}
(\mathbf{u}-\hat{\mathbf{p}}_i^r)
}{2}
\right).
\end{equation}

A reflection Gaussian contributes to a pixel only when its associated plane ID $\ell_i^r$ matches the plane ID $\mathbf{I}$ assigned to that pixel by TSDF raycasting. By substituting $\mathbf{x}_i^r$ with the reflection color $\mathbf{c}_i^r$ or scalar unity $1$, we obtain the reflection color image $\mathbf{C}_r=\mathcal{A}_r[\mathbf{c}^r]$ and reflection weight map $\mathbf{W}_r=\mathcal{A}_r[1]$, respectively.
Finally, the final rendered image is obtained by compositing the base image and reflection image using the reflection mask:
\begin{equation}
\mathbf{C}=\mathbf{C}_b+\mathbf{R}\cdot\mathbf{C}_r.
\label{equ:rendering}
\end{equation}

\section{Online Reconstruction Process}

As illustrated in Fig.~\ref{fig:pipeline}, our system takes an RGB-D stream as input and performs online reflection-aware reconstruction with a hybrid TSDF-Gaussian representation. The method is organized into two main parts: TSDF reconstruction and Gaussian reconstruction. In TSDF reconstruction, we preprocess each RGB-D frame, estimate camera poses using reflection-aware tracking, identify reflective planes from geometric, semantic, and temporal cues, and fuse observations into an augmented TSDF volume with reflection-aware attributes, as described in Sec.~\ref{sec:tsdf_reconstruction}. In Gaussian reconstruction, we incrementally add base and reflection Gaussians, followed by online optimization and pruning to maintain reconstruction quality and compactness, as described in Sec.~\ref{sec:gaussian_reconstruction}.

For real-time performance, tracking and TSDF fusion are performed for each incoming frame, while reflective plane identification and Gaussian optimization are triggered periodically at lower frequencies. This design enables stable tracking, reflection-aware map construction, and high-quality online reconstruction in reflective indoor scenes.

\subsection{TSDF reconstruction}
\label{sec:tsdf_reconstruction}

\subsubsection{Input preprocessing}
\label{sec:tracking}
Given the input color image $\mathbf{C}_{k}^{gt}$ and depth map $\mathbf{D}_k^{gt}$ at frame $k$, we first compute a local vertex map $\mathbf{V}_k^l$ and a local normal map $\mathbf{N}_k^l$ from the depth observation. Using the estimated camera pose, we then transform them into the global coordinate system, yielding a global vertex map $\mathbf{V}_k^g$ and a global normal map $\mathbf{N}_k^g$.

\subsubsection{Reflection-aware tracking}
In reflective scenes, view-dependent appearance caused by planar reflections violates photometric consistency and corrupts image features, making standard feature-based tracking unreliable. To address this, we adopt a two-stage reflection-aware tracking strategy.
In the first stage, an initial pose is estimated by aligning the current depth observation to the TSDF raycast surface using point-to-plane ICP:
\begin{equation}
E(\boldsymbol{\xi})=\sum\left\|\left(\mathbf{T}_{g, k}^{icp} \mathbf{V}_k^l(\mathbf{u})-\mathbf{V}_{k-1}^{*, g}(\hat{\mathbf{u}})\right) \cdot \mathbf{N}_{k-1}^{*, g}(\hat{\mathbf{u}})\right\|,
\end{equation}
where $\boldsymbol{\xi}$ is the Lie algebra representation of the transformation $\mathbf{T}_{g,k}^{icp}$, $\mathbf{u}$ denotes a pixel in the current frame, and $\hat{\mathbf{u}}=\text{proj}\left(\mathbf{K}\mathbf{T}_{k-1,k}\mathbf{V}_k^l(\mathbf{u})\right)$ is the projection of $\mathbf{V}_k^l(\mathbf{u})$ into the previous frame. Here, $\text{proj}(\cdot)$ denotes the projection operator, $\mathbf{V}_k^l(\mathbf{u})$ is the current vertex in the local camera frame, and $\mathbf{V}_{k-1}^{*,g}(\hat{\mathbf{u}})$ and $\mathbf{N}_{k-1}^{*,g}(\hat{\mathbf{u}})$ are the corresponding raycast vertex and normal in the global frame. Minimizing this energy yields an initial pose estimate $\mathbf{T}_{g,k}^{icp}$.

In the second stage, feature-based refinement is performed using ORB features, with the ICP-estimated $\mathbf{T}_{g,k}^{icp}$ as the initial pose. The reflection component is rendered from the current reflection Gaussians $\mathcal{G}_r$ at $\mathbf{T}_{g,k}^{\text{icp}}$ to obtain a reflection color image $\mathbf{C}_{k}^r$. Averaging its RGB channels and thresholding the intensity map at $0.1$ produces a binary tracking mask $\mathbf{R}_k^{\text{track}}$ that identifies pixels dominated by reflections. During ORB feature extraction~\cite{orbslam2}, only pixels with $\mathbf{R}_k^{\text{track}}(\mathbf{u})=0$ are considered, preventing unreliable reflective regions from corrupting tracking. The pose is then refined using a feature-based optimization backend adapted from ORB-SLAM2~\cite{orbslam2}, producing the final pose estimate $\mathbf{T}_{g,k}$.

This two-stage strategy combines the geometric stability of ICP with reflection-aware feature refinement, enabling robust tracking even in strongly reflective indoor scenes.

\subsubsection{Reflective plane identification}
\label{sec:plane_identification}
Reliable reflective plane detection is essential for reflection-aware tracking, rendering, and Gaussian reconstruction. In indoor scenes, planar geometry and semantic cues alone are insufficient: depth identifies planar structures, but cannot determine whether they produce reflections, while semantic segmentation highlights potentially reflective objects but cannot confirm reflection effects. To robustly identify reflective planes, we combine geometric, semantic, and temporal cues. Since semantic segmentation and planar analysis are computationally expensive, reflective plane identification is performed periodically every $N_p=10$ frames rather than for every incoming frame. 

Specifically, we first generate reflective plane candidates by combining geometric and semantic cues. We apply CAPE~\cite{cape}, a depth-based planar segmentation method, to the input depth map and extract a plane segmentation map. The detected local planes are then matched to the global plane set maintained by the system, producing a global plane ID map $\mathbf{I}_k^g$. In parallel, we apply Grounded SAM2~\cite{sam2,groundingdino} to the input color image to obtain a semantic segmentation map $\mathbf{Q}_k$. Taking advantage of its open-vocabulary capability, we predefine a set of object categories that are likely to exhibit specular reflections, such as floors, tables, and TVs. A geometric plane is regarded as a reflective-plane candidate if it is associated with one of these semantic categories. In our implementation, each geometric plane is associated with the semantic mask that has the largest overlap with the plane region.

We then validate these candidates using temporal color variance. During TSDF fusion, each voxel $\mathbf{p}$ maintains running statistics of voxel colors across the input sequence, and the corresponding variance is stored in $S_c(\mathbf{p})$ using Welford-style online updates~\cite{welford1962note}. At detection time, we raycast the TSDF to obtain a color variance map $\mathbf{C}_k^{var}$. For each candidate plane, we compute the fraction of image pixels associated with the plane whose temporal color variance exceeds the threshold $\delta_v=0.01$. If this fraction exceeds $\delta_{\text{area}}=0.05$, the plane is marked as reflective for the current detection step.

To improve robustness, we do not accept a reflective plane based on a single detection. Instead, for each plane, we maintain its reflective detection history over the most recent five windows. A plane is inserted into the reflective plane lookup table $L_{\mathrm{ref}}$ only if it is identified as reflective in more than three of these windows. After reflective planes are confirmed, we combine their associated semantic masks to form a predicted reflection mask for the current frame. This mask is then fused into the TSDF volume as a reflection-aware attribute, enabling the system to mark potential reflective regions for subsequent reconstruction and rendering.

\subsubsection{Augmented TSDF fusion}
\label{sec:tsdf_fusion}
We perform TSDF fusion to update voxel-wise SDF values and average colors. For frames without reflective plane identification, we only carry out standard TSDF fusion. When reflective plane identification is triggered at frame $k$, we additionally update reflection-aware voxel attributes. The color variance of each voxel, $S_c(\mathbf{p})$, is updated incrementally using a Welford-style online algorithm applied to the channel-averaged color from each new observation, capturing how much the voxel’s appearance varies over time. The plane ID is fused from the global plane map to preserve associations with planar structures, while the reflection strength $r(\mathbf{p})$ is updated using a sliding average over recent predicted reflection masks, providing a stable estimate of whether a voxel belongs to a reflective region. Together, these augmented attributes provide essential cues for subsequent reflection detection, reflection-aware tracking, and the initialization of reflection Gaussians. Detailed formulas and implementation specifics are provided in the supplementary material.

\subsection{Gaussian reconstruction}
\label{sec:gaussian_reconstruction}
\subsubsection{Gaussian densification}
We perform Gaussian densification separately for the base Gaussians $\mathcal{G}_b$ and the reflection Gaussians $\mathcal{G}_r$. In both cases, we first identify candidate pixels for adding Gaussians, then sample a subset of these pixels, and finally initialize new Gaussians from the sampled locations.
For notational clarity, at frame $k$, we denote the rendered color image by $\mathbf{C}_k$, the base Gaussian weight map by $\mathbf{W}_k^b$, and the reflection Gaussian weight map by $\mathbf{W}_k^r$.

For the base Gaussians $\mathcal{G}_b$, the pixels with high color error and low Gaussian coverage are selected as candidates for Gaussian adding:
\begin{equation}
\mathbf{M}_{\widetilde{\mathbf{u}}_b}=\{\widetilde{\mathbf{u}}_b| | \mathbf{C}_{k}^{gt}(\widetilde{\mathbf{u}}_b)-\mathbf{C}_k(\widetilde{\mathbf{u}}_b) \mid>\delta_{b} \text { and } W_{k}^{b}(\widetilde{\mathbf{u}}_b)<\delta_{w,b}\},
\end{equation}
where $\mathbf{C}_{k}^{gt}(\widetilde{\mathbf{u}}_b)$ and $\mathbf{C}_k(\widetilde{\mathbf{u}}_b)$ denote the input and rendered color images, used to compute the per-pixel color reconstruction error, and $W_k^b(\widetilde{\mathbf{u}}_b)$ is the base Gaussian weight at the pixel, indicating the current coverage by existing base Gaussians. Each candidate pixel is then backprojected into 3D using the observed depth to initialize the Gaussian position on the surface, while all remaining attributes—including scale, rotation, opacity, and SH coefficients—are initialized according to the standard rules of GPS-SLAM~\cite{gps-slam}.

For reflection Gaussians $\mathcal{G}_r$, candidate pixels are selected from regions marked as reflective that also exhibit high temporal color variance, large color reconstruction error, and low reflection Gaussian coverage:

\begin{equation}
\begin{aligned}
\mathbf{M}_{\widetilde{\mathbf{u}}_r}
=
\Big\{
\widetilde{\mathbf{u}}_r \ | \ 
\mathbf{R}_k(\widetilde{\mathbf{u}}_r)=1,\ 
|\mathbf{C}_{k}^{gt}(\widetilde{\mathbf{u}}_r)-\mathbf{C}_k(\widetilde{\mathbf{u}}_r)|>\delta_r, \\
\mathbf{C}_k^{var}(\widetilde{\mathbf{u}}_r)>\delta_v,\ 
\text{and } W_{k}^{r}(\widetilde{\mathbf{u}}_r)<\delta_{w,r}
\Big\},
\end{aligned}
\end{equation}
where $\mathbf{R}_k(\widetilde{\mathbf{u}}_r)$ denotes the binary reflection mask that identifies reflective regions, $\mathbf{C}_k^{var}(\widetilde{\mathbf{u}}_r)$ represents the color variance obtained by raycasting the TSDF, and $W_{k}^{r}(\widetilde{\mathbf{u}}_r)$ is the current reflection Gaussian weight, indicating the existing coverage at that pixel. Unlike base Gaussians, reflection Gaussians cannot be initialized directly from observed depth because they represent virtual content behind the reflective plane. For each sampled candidate pixel, the Gaussian position is initialized by uniformly sampling 10 points along the viewing ray starting from the ray-plane intersection and extending within a depth range of 2--8\,m. This strategy allows reflection Gaussians to represent virtual reflected content behind the reflective plane without requiring explicit reflected geometry. All remaining attributes are initialized in the same manner as for base Gaussians, and each newly added reflection Gaussian is assigned the corresponding plane ID.

\subsubsection{Gaussian optimization}
To preserve online efficiency, Gaussian optimization is triggered periodically every $N_o$ frames. After adding base and reflection Gaussians to the map, we perform online optimization using a subset of frames. The optimization uses $N_{local}$ frames evenly sampled from the observations collected since the previous optimization step to efficiently refine newly observed regions, along with $N_{global}$ randomly sampled historical keyframes to provide broader viewpoint coverage, which helps stabilize the optimization of reflection Gaussians across viewpoints. The keyframe set is maintained using a standard motion-based heuristic: a new keyframe is inserted when either the relative rotation to the last keyframe exceeds $\delta_{\mathrm{angle}}$ or the relative translation exceeds $\delta_{\mathrm{move}}$.

We jointly optimize the Gaussian groups $\mathcal{G}_b$ and $\mathcal{G}_r$ by minimizing the following loss function:
\begin{equation}
\mathcal{L} = \mathcal{L}_{color} + \lambda_d \mathcal{L}_{depth} + \lambda_g \mathcal{L}_{grad},
\end{equation}
where the regularization term is defined as
\begin{equation}
\quad \mathcal{L}_{grad} = |\nabla \mathbf{C}_r|_1.
\end{equation}

Here, the color term $\mathcal{L}_{color}$ combines an $L_1$ loss and an SSIM loss between the rendered image $\mathbf{C}$ and the input RGB image $\mathbf{C}_{gt}$, while the depth term $\mathcal{L}_{depth}$ uses an $L_1$ loss between the rendered depth $\mathbf{D}_b$ and the input depth $\mathbf{D}_{gt}$. We set the depth weight $\lambda_d$ to $0.2$ and the reflection smoothness weight $\lambda_g$ to $0.05$ in all experiments. The smoothness term $\mathcal{L}_{grad}$, computed via the Sobel operator $\nabla$, encourages spatially coherent reflection image $\mathbf{C}_r$, which facilitates the decomposition of diffuse and reflective components in practice.

\subsubsection{Gaussian pruning}

After each optimization round, we prune both base and reflection Gaussians according to their scale and opacity. Gaussians with excessively large scales tend to oversmooth local structures, while those with very small scales or low opacity contribute little to rendering and often correspond to unstable or redundant primitives. Removing such Gaussians helps maintain a compact and stable representation during long-term online reconstruction. Detailed threshold settings are provided in the supplementary material.

\section{Experiments}

\begin{figure*}[h]
    \centering
    \includegraphics[width= \textwidth]{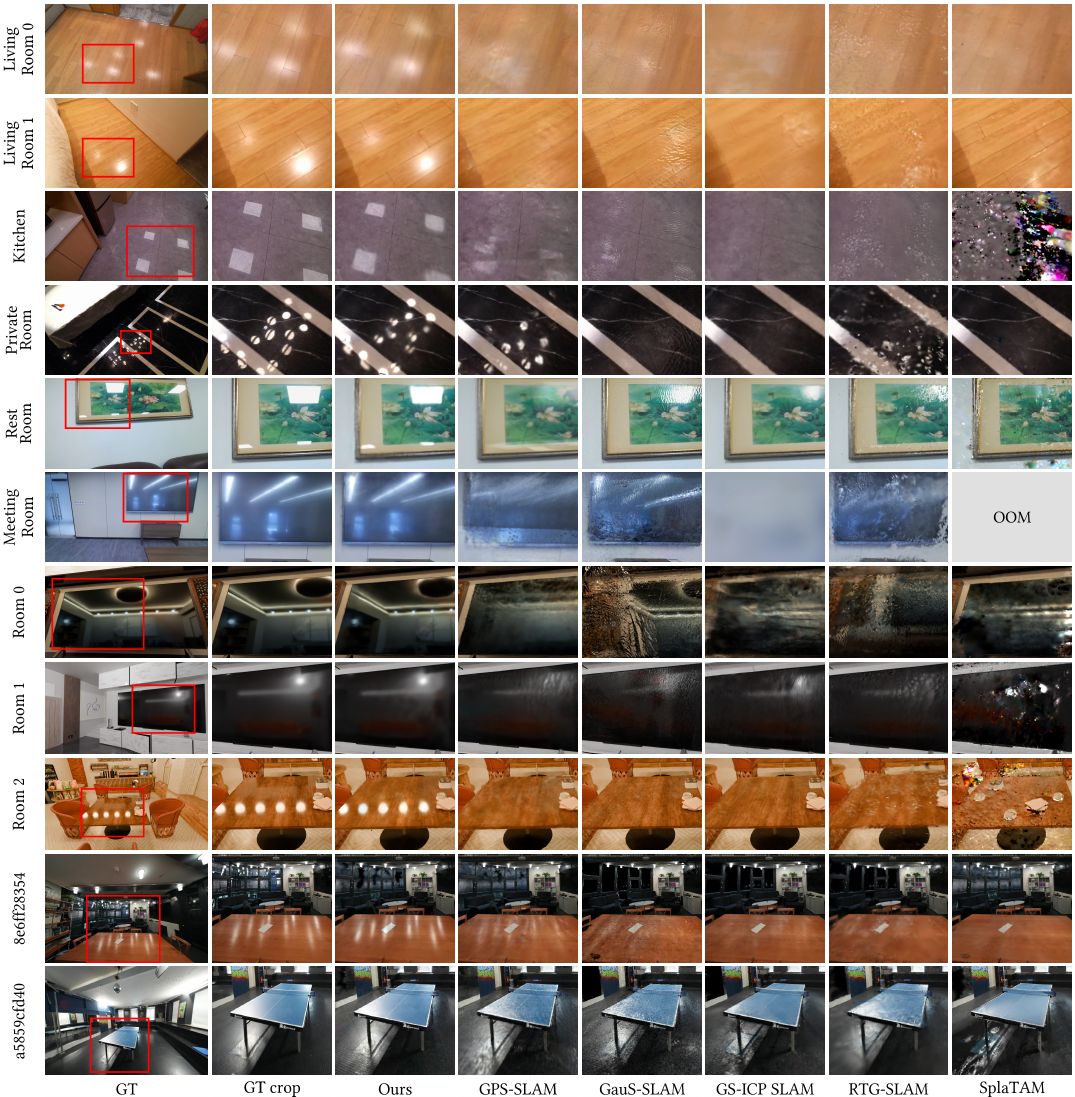}
    \Description{
    A qualitative comparison figure of rendering quality results on reflective indoor scenes. Each row shows one scene with the ground-truth image, a cropped reflective region, and rendering results from several baseline methods and the proposed method. Red boxes highlight regions with reflection-related artifacts, such as blurry reflections, noisy reflective surfaces, drift, or missing reflective content. The proposed method produces sharper and more stable reflections compared with the baselines.
    }
    \caption{Qualitative comparison of rendering quality across multiple reflective indoor scenes. The first two columns show the original GT images and the cropped regions for clearer visualization of reflective areas. Our method consistently produces sharper and more stable reflections. Red boxes indicate regions where baseline methods (GPS-SLAM, GauS-SLAM, GS-ICP SLAM, RTG-SLAM, and SplaTAM) suffer from drifting or fail to reconstruct reflections accurately. Note that SplaTAM runs out of memory (OOM) on \textit{Meeting Room}, so no result is reported for it.}
    \label{fig:rq_cmp}
\end{figure*}

\begin{figure*}[h]
    \centering
    \includegraphics[width= \textwidth]{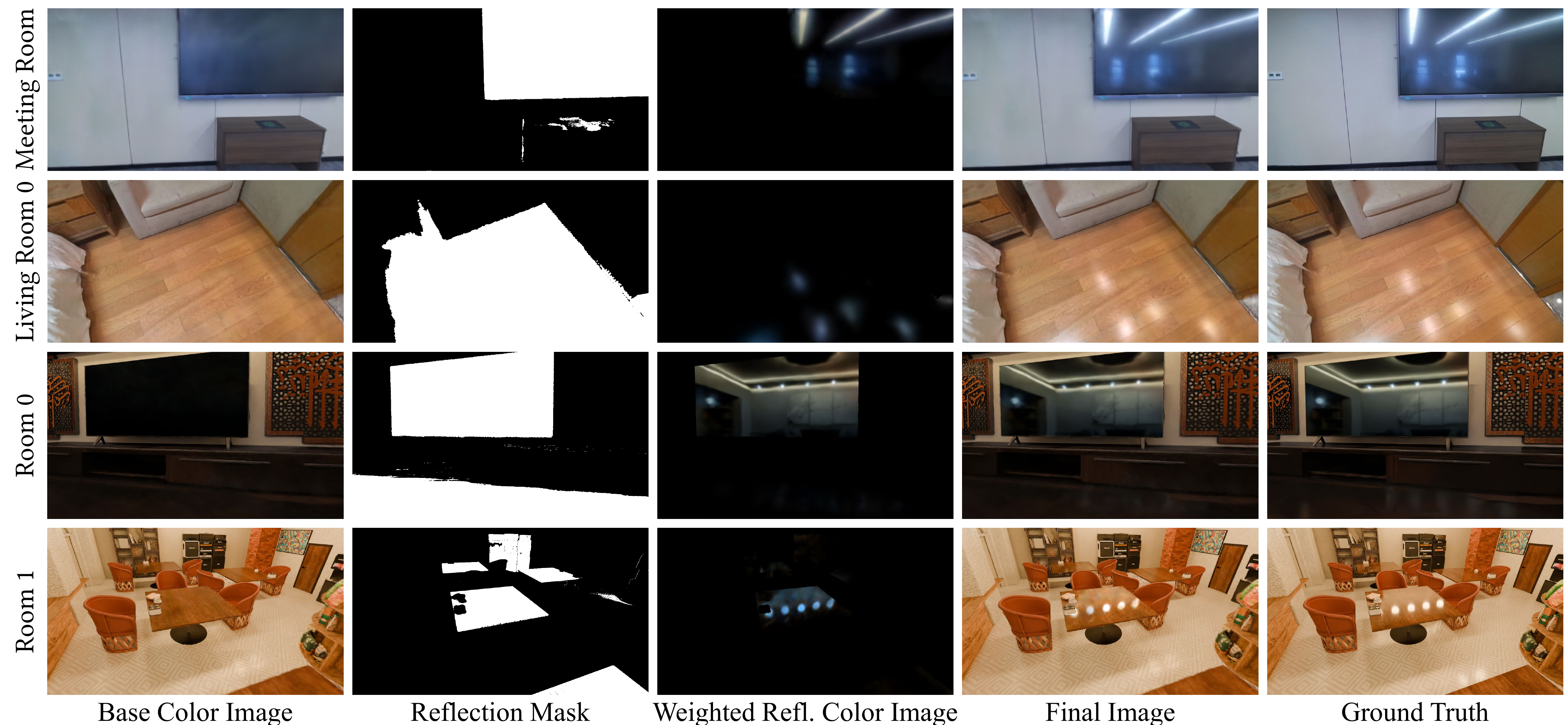}
    \caption{
    Visualization of rendering components on four representative scenes. From left to right, we show the base color image, reflection mask, weighted reflection color image, final rendered image, and ground-truth image. The base color mainly captures diffuse appearance, while the reflection mask localizes reflective regions and the weighted reflection color image shows the contribution of reflection Gaussians. The final image combines the base and reflection components to better reproduce reflective appearance.
    }
    \Description{
    A visualization figure showing rendering decomposition results on four scenes. Each row corresponds to one scene, and the columns show the base color image, reflection mask, weighted reflection color image, final rendered image, and ground-truth image. The figure illustrates how the proposed method separates diffuse base appearance from reflective components and combines them to produce the final rendering.
    }
    \label{fig:rq_decom}
\end{figure*}

\subsection{Experiment setup}
\paragraph{Implementation Details}
The proposed SLAM system is implemented and evaluated on a desktop equipped with an AMD 9950X3D CPU and an NVIDIA RTX 4090 GPU.
The core SLAM framework is implemented in C++ based on GPS-SLAM~\cite{gps-slam}. We develop custom CUDA kernels for rasterization and backpropagation. The CAPE~\cite{cape} plane detection module is also implemented in C++ by modifying the original CAPE repository.
In addition, the Grounded SAM2~\cite{groundingdino,sam2} segmentation module is implemented in Python and communicates with the C++ SLAM system through shared memory for efficient data exchange. For more details, please refer to the supplementary material.

\paragraph{Datasets}
We primarily evaluate our method on a self-constructed dataset with noticeable reflection effects, denoted as RIRD (Reflection Indoor RGB-D). RIRD consists of six real indoor scenes captured with an Azure Kinect RGB-D camera and three synthetic reflective indoor scenes rendered using Blender. We refer to the real subset as RIRD-Real and the synthetic subset as RIRD-Syn. RIRD-Real is mainly used for qualitative evaluation of novel-view rendering and reconstruction quality, while RIRD-Syn is used to quantitatively evaluate tracking and geometric reconstruction accuracy in reflective environments.

We also evaluate our method on two reflective ScanNet++~\cite{yeshwanth2023scannet++} scenes to assess novel-view rendering and geometric reconstruction in real-world settings. The selected scenes are \texttt{a5859cfd40} and \texttt{8e6ff28354}.
In addition, we use Replica~\cite{straub2019replica} and TUM-RGBD~\cite{tum} as standard RGB-D benchmarks. Replica provides synthetic indoor scenes without noticeable reflections, while TUM-RGBD provides real RGB-D sequences with accurate ground-truth trajectories for tracking evaluation.

\paragraph{Baselines}
We compare our method with several state-of-the-art 3DGS-based RGB-D SLAM systems, including SplaTAM~\cite{splatam}, RTG-SLAM~\cite{rtg-slam}, GauS-SLAM~\cite{gaus-slam}, GS-ICP SLAM~\cite{gs-icp}, and GPS-SLAM~\cite{gps-slam}.
SplaTAM employs a Gaussian-based tracking and mapping pipeline via differentiable rendering. RTG-SLAM proposes a compact Gaussian representation to significantly reduce memory usage and computational cost. GauS-SLAM adopts a 2D Gaussian surfel-based representation for robust tracking and accurate dense reconstruction. GS-ICP SLAM combines Gaussian-based scene representation with ICP-based tracking to achieve real-time pose estimation. GPS-SLAM introduces a hybrid Gaussian-SDF representation and enables ultra-fast RGB-D SLAM. In addition, we include ORB-SLAM2~\cite{orbslam2} and BundleFusion~\cite{bundlefusion} as representatives of classical SLAM systems to evaluate tracking performance in reflective indoor environments. We reproduce the results using their official implementations and run all experiments on the same computer.

\begin{figure*}[t]
    \centering
    \includegraphics[width= \textwidth]{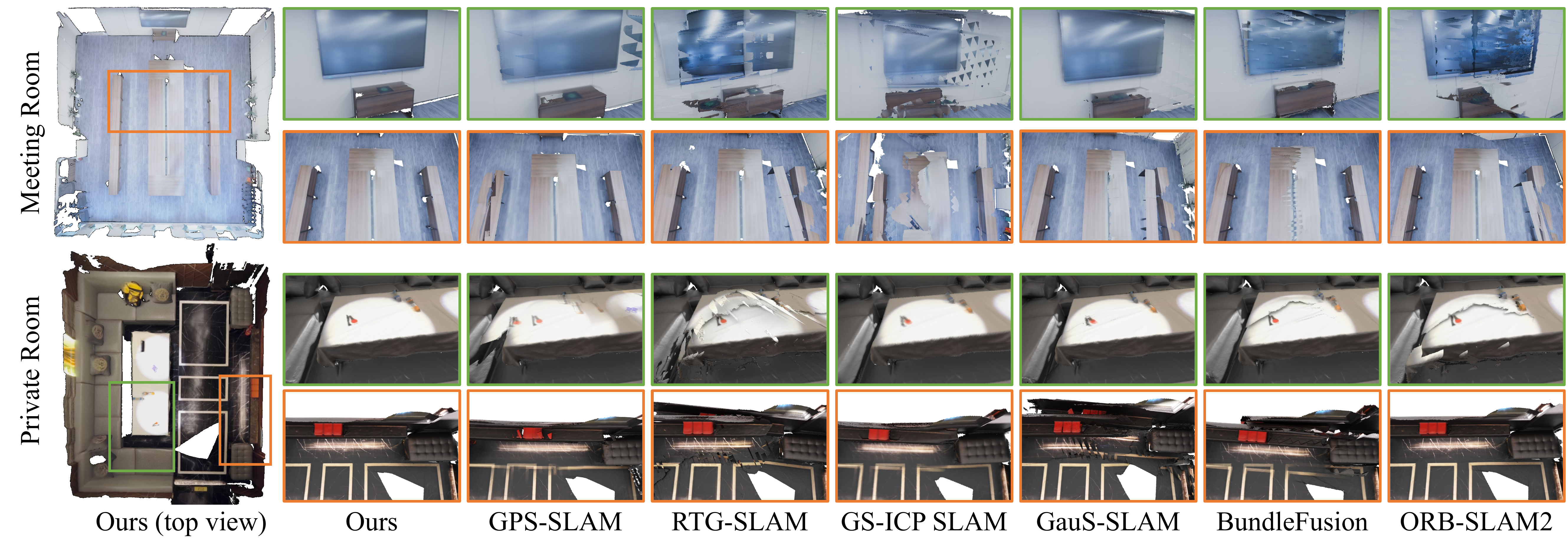}
    \caption{Comparison of reconstructed meshes in reflective indoor scenes (\textit{Meeting Room} and \textit{Private Room}). Our method produces more complete and accurate reconstructions in regions containing a glossy tiled floor and a TV screen. Baseline methods, including GPS-SLAM~\cite{gps-slam}, RTG-SLAM~\cite{rtg-slam}, GS-ICP SLAM~\cite{gs-icp}, GauS-SLAM~\cite{gaus-slam}, BundleFusion~\cite{bundlefusion}, and ORB-SLAM2~\cite{orbslam2}, exhibit drifting or misalignment in these reflective regions.}
    \Description{
    A qualitative comparison figure of reconstructed meshes in two reflective indoor scenes, Meeting Room and Private Room. The figure compares the proposed method with several baseline SLAM methods. Reflective regions such as glossy tiled floors and TV screens are highlighted. The baseline methods show artifacts such as drift, misalignment, incomplete geometry, or blurred reconstruction, while the proposed method produces cleaner and more consistent mesh reconstruction in reflective regions.
    }
    \label{fig:mesh_recon}
\end{figure*}

\subsection{Evaluation}

\subsubsection{Rendering quality} We evaluate rendering quality with a particular focus on reflection-related appearance. For novel-view synthesis, we conduct experiments on the synthetic subset of RIRD and on ScanNet++. In each RIRD-Syn scene, we uniformly sample 25 viewpoints and perturb their poses to render novel-view images, while for ScanNet++ we select one test view every eight frames. Input-view rendering is evaluated on the real subset of RIRD and Replica, following the protocols used in prior Gaussian-based SLAM methods~\cite{splatam,rtg-slam,gs-icp}.

Quantitative results are summarized in Table~\ref{table:rq_cmp}. On the reflective datasets, our method achieves the best novel-view synthesis quality on RIRD-Syn and ScanNet++, and also attains the best input-view rendering results on RIRD-Real. On Replica, which does not contain noticeable reflections, our method achieves rendering quality comparable to other methods. These results show that our reflection-aware representation improves rendering quality in reflective scenes while remaining compatible with general indoor scenes.

\begin{table}[t]
\caption{
Rendering quality comparison on RIRD-Syn, ScanNet++, RIRD-Real, and Replica. We report PSNR, SSIM, and LPIPS. RIRD-Syn and ScanNet++ are evaluated under the novel-view setting, while RIRD-Real and Replica are evaluated on training views. RIRD-Syn, ScanNet++, and RIRD-Real contain reflective indoor scenes, whereas Replica is used to evaluate performance on general indoor scenes without reflections.
}
\Description{
A quantitative table comparing rendering quality across four datasets: RIRD-Syn, ScanNet++, RIRD-Real, and Replica. The rows list different SLAM methods, and each method is evaluated using PSNR, SSIM, and LPIPS. The proposed method achieves the best performance on the reflective datasets and remains competitive on Replica.
}
\footnotesize
\resizebox{0.95\columnwidth}{!}{
\begin{tabular}{c|c|cccc}
\toprule
\multicolumn{1}{c|}{Method}  & \multicolumn{1}{c|}{Metric} & RIRD-Syn & ScanNet++               & RIRD-Real               & \multicolumn{1}{l}{Replica} \\ \hline
\multirow{3}{*}{SplaTAM}     & PSNR$\uparrow$              & 18.66             & 23.85                   & 22.69                   & 34.14                       \\
                             & SSIM$\uparrow$              & 0.634             & 0.834                   & 0.820                   & 0.936                       \\
                             & LPIPS$\downarrow$           & 0.422             & 0.241                   & 0.355                   & 0.142                       \\ \hline
\multirow{3}{*}{RTG-SLAM}    & PSNR$\uparrow$              & 17.81             & 22.37                   & \cs{25.16}              & 34.04                       \\
                             & SSIM$\uparrow$              & 0.650             & 0.783                   & 0.855                   & 0.924                       \\
                             & LPIPS$\downarrow$           & 0.444             & 0.299                   & 0.359                   & 0.183                       \\ \hline
\multirow{3}{*}{GauS-SLAM}   & PSNR$\uparrow$              & 17.02             & \cs{25.17}              & 24.77                   & \cf{38.85}     \\
                             & SSIM$\uparrow$              & 0.497             & 0.800                   & 0.815                   & \cf{0.974}     \\
                             & LPIPS$\downarrow$           & 0.428             & 0.287                   & 0.330                   & \cf{0.068}     \\ \hline
\multirow{3}{*}{GS-ICP SLAM} & PSNR$\uparrow$              & 22.70             & 24.73                   & 24.93                   & 33.00                       \\
                             & SSIM$\uparrow$              & 0.798             & \cs{0.867}              & \cs{0.864} & 0.926                       \\
                             & LPIPS$\downarrow$           & 0.272             & \cs{0.194}              & \cs{0.289} & 0.116                       \\ \hline
\multirow{3}{*}{GPS-SLAM}    & PSNR$\uparrow$              & \cs{26.79}        & 24.59                   & 24.46      & 37.25                       \\
                             & SSIM$\uparrow$              & \cs{0.871}        & 0.836                   & 0.849                   & 0.955                       \\
                             & LPIPS$\downarrow$           & \cs{0.208}        & 0.255                   & 0.322                   & 0.108                       \\ \hline
\multirow{3}{*}{Ours}        & PSNR$\uparrow$              & \cf{30.29}        & \cf{26.01}              & \cf{30.82} & \cs{37.29}     \\
                             & SSIM$\uparrow$              & \cf{0.909}        & \cf{0.869}              & \cf{0.921} & \cs{0.961}     \\
                             & LPIPS$\downarrow$           & \cf{0.191}        & \cf{0.185}              & \cf{0.218} & \cs{0.102}     \\ 
\bottomrule
\end{tabular}
}
\label{table:rq_cmp}
\end{table}

\begin{figure*}[t]
    \centering
    \includegraphics[width=0.9\textwidth,height=0.9\textheight,keepaspectratio]{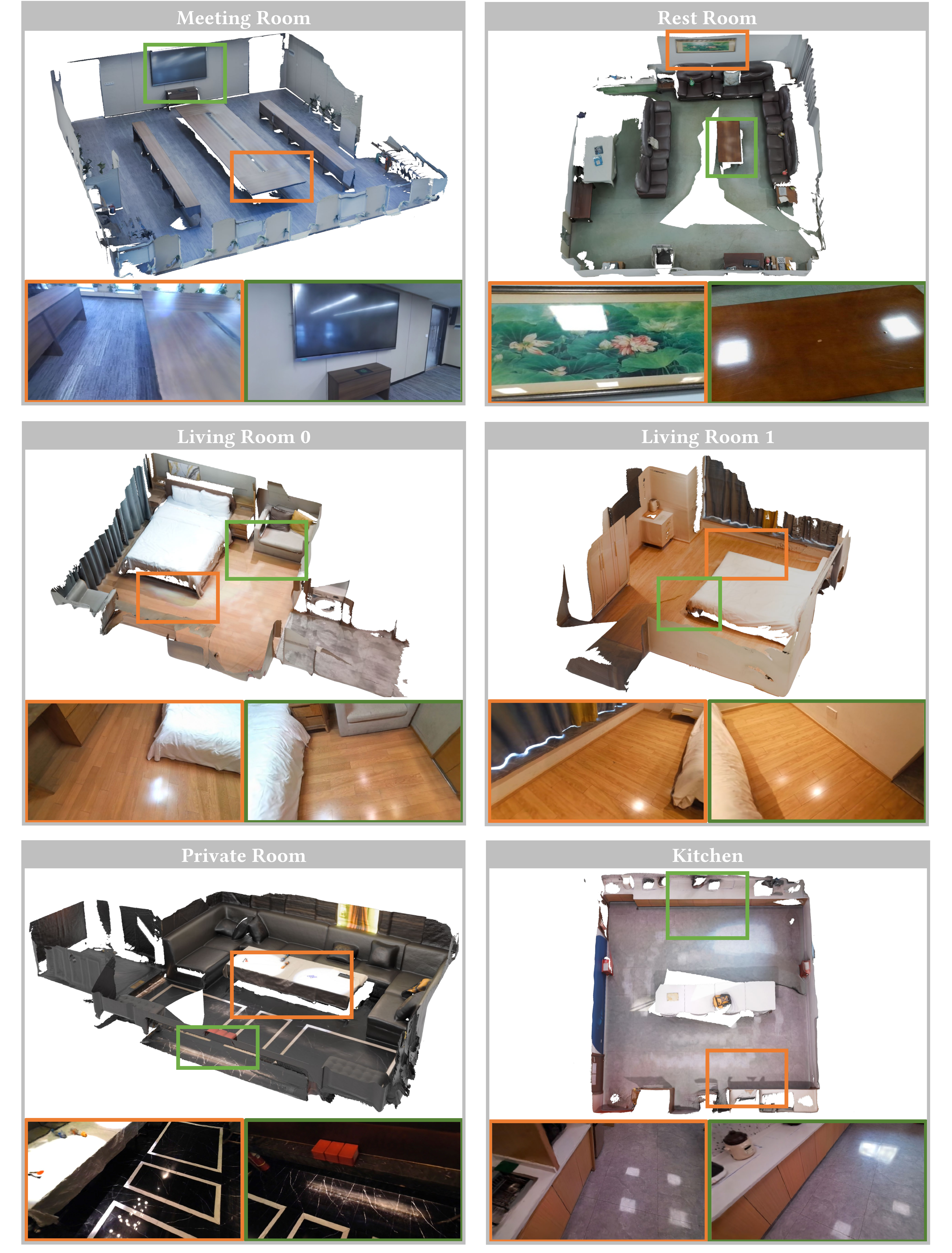}
    \caption{
    Reconstruction and novel-view synthesis results on six real RIRD-Real scenes. For each scene, we show the reconstructed scene overview and zoomed-in views of reflective regions. Colored boxes link the overview to the corresponding zoomed-in renderings. Our method produces coherent scene-scale reconstruction and visually consistent reflective appearance in real reflective indoor environments.
    }
    \Description{
    A qualitative figure showing reconstruction and novel-view synthesis results on six real indoor scenes from RIRD-Real, including Meeting Room, Rest Room, Living Room 0, Living Room 1, Private Room, and Kitchen. Each scene contains a reconstructed scene overview and two zoomed-in views of reflection-rich regions. Colored boxes link regions in the reconstructed overview to the corresponding zoomed-in renderings. The figure illustrates that the proposed method reconstructs complete scene layouts and preserves reflective appearance on surfaces such as floors, tables, TV screens, and wall paintings.
    }
    \label{fig:mesh_topview}
\end{figure*}

We present qualitative comparisons on RIRD-Real, RIRD-Syn, and ScanNet++ in Fig.~\ref{fig:rq_cmp}. Existing methods do not explicitly model reflective appearance, and as a result, their reconstructions exhibit artifacts in reflective regions. On surfaces with relatively high roughness, existing methods often produce over-smoothed textures, as seen on the wooden floors in the \textit{Living Room 0} and \textit{Living Room 1} scenes. On smoother surfaces with stronger and clearer reflections, the rendered appearance is often noisy, as observed on the tiled floor in the \textit{Kitchen} scene reconstructed by SplaTAM, and on the TV screen in the \textit{Meeting Room} scene reconstructed by GauS-SLAM and RTG-SLAM. In contrast, our method better models reflective appearance across surfaces with varying roughness, producing more coherent and visually plausible results. We further illustrate our reconstruction and novel-view synthesis results on the real scenes captured in RIRD in Fig.~\ref{fig:mesh_topview}. To better visualize the high-quality reflective rendering produced by our method, we refer readers to the supplementary video.

\paragraph{Reflection decomposition.}
To explain the improved rendering quality over the baselines, we visualize the decomposed rendering components in Fig.~\ref{fig:rq_decom}. For each example, we show the base color image $\mathbf{C}_b$, the reflection mask $\mathbf{R}$, the weighted reflection color $\mathbf{R} \cdot \mathbf{C}_r$, the final composited result $\mathbf{C}$, and the ground truth image $\mathbf{C}_{gt}$. The base color mainly captures diffuse appearance, while the reflection color focuses on view-dependent reflective content associated with planar surfaces. By separating planar reflections from diffuse appearance, our method avoids mixing reflective content with the base scene and therefore produces cleaner and more accurate renderings.


\subsubsection{Tracking accuracy} We report absolute trajectory error (ATE) on RIRD-Syn, Replica, and TUM-RGBD, as shown in Table~\ref{table:ate}. Our method achieves the lowest ATE on the reflective RIRD-Syn dataset, demonstrating its robustness to reflection-induced tracking interference. On Replica and TUM-RGBD, our method also performs competitively, indicating that the proposed reflection-aware tracking strategy remains compatible with general indoor scenes.

\begin{table}[t]
\caption{
Tracking accuracy comparison on RIRD-Syn, Replica, and TUM-RGBD. We report the absolute trajectory error (ATE), where lower values indicate better tracking accuracy. RIRD-Syn contains reflective synthetic scenes, while Replica and TUM-RGBD are used to evaluate tracking performance on general indoor scenes.
}
\Description{
A quantitative table comparing tracking accuracy across RIRD-Syn, Replica, and TUM-RGBD. The rows list different SLAM methods, and the columns report absolute trajectory error for each dataset. The proposed method achieves the lowest error on RIRD-Syn and TUM-RGBD, and competitive performance on Replica.
}

\begin{tabular}{c|ccc}
\toprule
 Method      & RIRD-Syn  & Replica   & TUM  \\ \hline
 ORB-SLAM2   &  22.26    & 0.36      & 1.27   \\ 
 BundleFusion &  8.52    & 0.42      & 1.53     \\
 SplaTAM     & 36.15     & 0.39      & 1.81 \\
 RTG-SLAM    & 10.71     & 0.19      & \cs{1.13} \\
 GauS-SLAM   & 41.81     & \cf{0.07} & 1.42 \\
 GS-ICP SLAM & 16.62     & 0.18      & 2.33 \\
 GPS-SLAM    & \cs{0.69} & 0.19      & 2.75 \\
 Ours        & \cf{0.31} & \cs{0.17} & \cf{1.06}\\
\bottomrule
\end{tabular}
\label{table:ate}
\end{table}


\paragraph{Mesh reconstruction under estimated trajectories.}
To more clearly illustrate the improvement in tracking performance under reflective conditions, we select two real scenes from RIRD-Real and visualize reconstructed meshes, as shown in Fig.~\ref{fig:mesh_recon}. For a fair comparison, we reconstruct a mesh for each method using TSDF fusion with its estimated poses and input depth maps. In the \textit{Meeting Room} scene, GPS-SLAM and GS-ICP SLAM rely on ICP-based geometric alignment and exhibit noticeable drift in this large-scale environment, especially in regions with sparse geometric features. Classical methods such as BundleFusion and ORB-SLAM2 also exhibit noticeable drift around the TV region in \textit{Meeting Room} due to unreliable photometric cues. RTG-SLAM, which employs ORB-SLAM2 as its back-end, shows similar drift patterns. GauS-SLAM uses both color and depth constraints for pose optimization, so view-dependent reflections can disturb the photometric term and lead to pose misalignment. In contrast, our reflection-aware method combines masked ORB feature refinement with ICP initialization, effectively suppressing reflection-induced interference and producing more stable and accurate tracking in reflective regions.

\begin{figure*}[t]
    \centering
    \includegraphics[width= \textwidth]{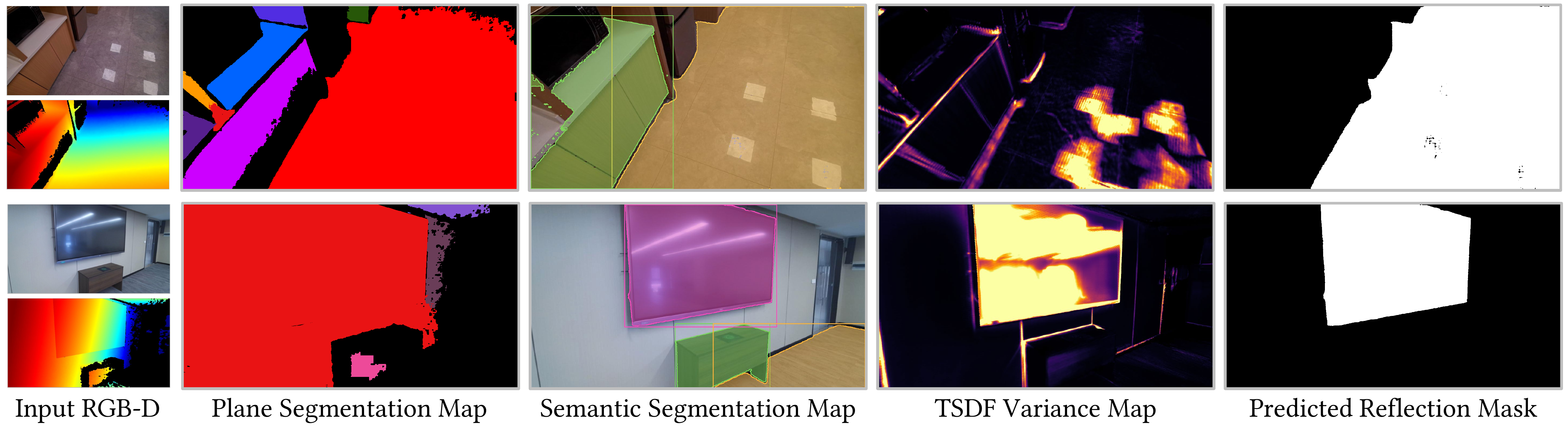}
    \caption{
    Visualization of reflective plane identification. From left to right, we show the input RGB-D frame, plane segmentation map, semantic segmentation map, TSDF variance map, and predicted reflection mask. Different colors in the plane segmentation map indicate different planar regions. In the semantic segmentation map, TV \protect\textcolor{magenta}{$\blacksquare$}, table \protect\textcolor{green}{$\blacksquare$}, and floor \protect\textcolor{yellow!90!green}{$\blacksquare$} indicate the semantic categories. The TSDF variance map is colorized according to temporal color variance, where brighter colors indicate higher variance. By combining geometric, semantic, and temporal cues, our method predicts reliable reflective regions for reflection-aware reconstruction.
    }
    \Description{
    An overview figure showing the reflective plane identification process. The figure contains five columns: an input RGB-D frame, a plane segmentation map with different planar regions shown in different colors, a semantic segmentation map highlighting reflection-prone categories such as TV, table, and floor, a TSDF temporal color variance map where brighter regions indicate higher variance, and the final predicted reflection mask. The figure illustrates how geometric plane segmentation, semantic cues, and temporal variance are combined to identify reflective regions.
    }                     
    \label{fig:plane_detect}
\end{figure*}

\subsubsection{Geometry quality} 
Following NICE-SLAM~\cite{nice-slam}, we evaluate geometric reconstruction quality on ScanNet++ and RIRD-Syn using the metrics of Accuracy, Completion, Accuracy Ratio (<3 cm), and Completion Ratio (<3 cm). Since ScanNet++ is not designed for SLAM and exhibits abrupt changes between consecutive frames, we use ground-truth poses for mapping in experiments on this dataset.
For each method, we extract the geometry following the strategy used in its original paper.
For SplaTAM, RTG-SLAM, and GS-ICP SLAM, we randomly sample a fixed number of Gaussian points for evaluation. For GauS-SLAM, we follow the original implementation and reconstruct a scene mesh by applying TSDF fusion to the rendered depth maps. For GPS-SLAM and our method, we reconstruct the scene mesh by extracting surfaces from the TSDF volume using marching cubes. On ScanNet++, with provided camera poses, our method achieves geometry quality on par with existing Gaussian-based SLAM methods.
On RIRD-Syn, our method shows a clear advantage over the other methods. We attribute this improvement mainly to our more accurate tracking.

\begin{table}[t]
\caption{
Geometry reconstruction quality comparison on RIRD-Syn and ScanNet++. We report accuracy, completion, accuracy ratio, and completion ratio. The ratio metrics measure the percentage of points with errors below 3 cm.
}
\Description{
A quantitative table comparing geometry reconstruction quality on RIRD-Syn and ScanNet++. The rows list different SLAM methods, and each method is evaluated using accuracy, accuracy ratio, completion, and completion ratio. The proposed method achieves the best geometry quality on RIRD-Syn and competitive results on ScanNet++.
}
\footnotesize
\resizebox{\columnwidth}{!}{
\begin{tabular}{c|c|cccc}
\toprule
Method                       & Dataset   & Acc$\downarrow$        & Acc Ratio$\uparrow$     & Com$\downarrow$        & Com Ratio$\uparrow$     \\ \hline
\multirow{2}{*}{SplaTAM}     & ScanNet++ & 3.10                   & 65.55                   & 3.03                   & 61.21                   \\
                             & RIRD-Syn  & 6.32                   & 44.64                   & 4.74                   & 50.52                   \\ 
                             \hline
\multirow{2}{*}{RTG-SLAM}    & ScanNet++ & 1.41                   & 91.84                   & 1.74                   & 87.18                   \\
                             & RIRD-Syn  & 7.85                   & 62.69                   & 5.77                   & 59.09                        \\
                             \hline
\multirow{2}{*}{GauS-SLAM}   & ScanNet++ & 3.92                   & 74.13                   & 1.49                   & 95.84                   \\
                             & RIRD-Syn  & 60.01                  & 63.22                   & 45.58                  & 64.22                   \\
                             \hline
\multirow{2}{*}{GS-ICP SLAM} & ScanNet++ & \cf{0.87}              & \cf{99.94}              & \cf{1.09}              & 96.76                   \\
                             & RIRD-Syn  & 4.93                   & 84.69                   & 5.12                   & 75.96                   \\
                             \hline
\multirow{2}{*}{GPS-SLAM}    & ScanNet++ & \cs{1.24}              & \cs{95.21}              & 1.11                   & \cs{98.50}               \\
                             & RIRD-Syn  & \cs{3.4}                    & \cs{90.25}                   & \cs{1.90}                    & \cs{94.35}                   \\
                             \hline
\multirow{2}{*}{Ours}        & ScanNet++ & 1.25                   & 95.18                   & \cs{1.10}              & \cf{98.53}                \\
                             & RIRD-Syn  & \cf{0.95}              & \cf{94.74}              & \cf{0.96}              & \cf{97.56}                    \\ \bottomrule
\end{tabular}
}
\end{table}

\subsubsection{Efficiency analysis} Table~\ref{table:performanceshort} summarizes the runtime, memory usage, Gaussian count, and rendering quality of all methods on the \textit{Office 0} scene from Replica and the \textit{Meeting Room} scene from RIRD-Real. On the Replica \textit{Office 0} scene, our method incurs only a negligible performance overhead compared with GPS-SLAM, while achieving comparable rendering quality. This indicates that the proposed reflection-aware representation remains efficient in general indoor scenes without noticeable reflections. On the reflective \textit{Meeting Room} scene, GPS-SLAM inserts substantially more Gaussians, reaching 2.37M primitives. As a result, its runtime drops to 51.0 FPS. In contrast, although our method jointly optimizes both base and reflection Gaussians, the proposed representation models reflective appearance more effectively. It requires only 0.64M Gaussians and achieves the highest frame rate of 70.5 FPS, while also producing substantially higher rendering quality.

\begin{table}[h]
\caption{
Runtime, memory, and rendering quality comparison on Replica \textit{Office 0} and RIRD-Real \textit{Meeting Room}. We report mapping time per frame, FPS, Gaussian count, GPU memory usage, and PSNR. Replica \textit{Office 0} represents a general indoor scene, while \textit{Meeting Room} contains strong planar reflections.
}
\Description{
A quantitative table comparing runtime, memory usage, Gaussian count, and PSNR across different SLAM methods on Replica office 0 and the RIRD-Real Meeting Room scene. The proposed method achieves high runtime and compact Gaussian count, especially on the reflective Meeting Room scene, while maintaining the best PSNR among the compared methods.
}
\footnotesize
\resizebox{\columnwidth}{!}{

\begin{tabular}{c|c|ccccc}
\toprule
Method                         & Dataset & \makecell{Mapping\\/Frame (ms)} & FPS    & \makecell{Gaussian\\ Count}           & Memory (MB)                                     & PSNR  \\
\hline
\multirow{2}{*}{SplaTAM}       & Replica & 3066.7        & 0.3         & 6.22M      & 9551      & 38.18 \\
                               & RIRD-Real   & --        & --         & --      & OOM      & -- \\ 
                               \hline
\multirow{2}{*}{RTG-SLAM}      & Replica & 59.9          & 16.6        & 0.82M       & \cf{2755} & 37.78 \\
                               & RIRD-Real   & 82.4            & 12.12        & \cs{1M}  & \cs{10656}  & \cs{22.42} \\ \hline
\multirow{2}{*}{GauS-SLAM}     & Replica & 314.9         & 3.1         & 6.69M      & 15745     & \cf{42.40} \\
                               & RIRD-Real   & 376.1         & 2.6         & 10.05M      & 23891     & 17.89  \\
                               \hline
\multirow{2}{*}{GS-ICP SLAM}   & Replica & 5.4           & 184.8       & 1.72M      & 3677      & 37.79 \\
                               & RIRD-Real   & \cs{16.7}          & \cs{59.8}        & 2.41M      & \cf{7528}      & 13.53 \\ \hline
\multirow{2}{*}{GPS-SLAM}      & Replica & \cf{2.4}      & \cf{408.1}  & \cs{0.11M}  & \cs{3603} & 40.91 \\
                               & RIRD-Real   & 19.5      & 51.0  & 2.37M       & 17719 & 18.42 \\
                               \hline
\multirow{2}{*}{Ours}          & Replica & \cs{2.8}      & \cs{358.8}   & \cf{0.10M}  & 5714      & \cs{41.30} \\
                               & RIRD-Real    & \cf{14.18}     & \cf{70.5}      & \cf{0.64M}  & 14017      & \cf{28.35} \\
\bottomrule
\end{tabular}
}
\label{table:performanceshort}
\end{table}

\paragraph{Runtime breakdown.}
Table~\ref{tab:performance_detail} reports the per-frame runtime of major components on the RIRD-Real \textit{Meeting Room} scene. Semantic segmentation is the most expensive pre-mapping module, taking 7.04 ms per frame. The total pre-mapping runtime, including pose tracking, plane detection, semantic segmentation, TSDF fusion, and other preprocessing overhead, is 10.12 ms, which is still lower than the 14.18 ms used by Gaussian mapping. This indicates that Gaussian mapping remains the main runtime bottleneck. Since these pre-mapping modules can run in parallel with Gaussian mapping, they have limited impact on the system speed.

\begin{table}[h]
\caption{
Runtime breakdown on the RIRD-Real \textit{Meeting Room} scene. We report the average per-frame runtime of pose tracking, plane detection, semantic segmentation, TSDF fusion, the total pre-mapping stage, and Gaussian mapping.
}
\Description{
A table reporting the per-frame runtime of major components on the RIRD-Real Meeting Room scene. Pose tracking takes 0.85 ms, plane detection takes 1.88 ms, semantic segmentation takes 7.04 ms, TSDF fusion takes 0.19 ms, the total pre-mapping stage including additional lightweight overhead takes 10.12 ms, and Gaussian mapping takes 14.18 ms.
}
\footnotesize
\begin{tabular}{l|c}
\toprule
Module                        & Time / Frame (ms) \\
\hline
Pose Tracking                 & 0.85              \\
Plane Detection               & 1.88              \\
Semantic Segmentation         & 7.04              \\
TSDF Fusion                   & 0.19              \\
Pre-mapping Stage             & 10.12             \\
Gaussian Mapping              & 14.18             \\
\bottomrule
\end{tabular}
\label{tab:performance_detail}
\end{table}

\paragraph{Rendering speed.}
Table~\ref{tab:render_speed} reports the per-pass rendering time and the number of Gaussians used in each pass on the RIRD-Real \textit{Meeting Room} scene. Despite using three rendering passes, our reflection-aware representation keeps the total number of Gaussians moderate, with 0.30M base Gaussians and 0.34M reflection Gaussians in this scene. As a result, the combined rendering time of all passes is only 3.22 ms per frame, corresponding to 310.0 FPS. This demonstrates that the proposed decomposition enables efficient novel-view synthesis without introducing excessive Gaussian redundancy.

\begin{table}[h]
\caption{
Rendering performance breakdown on the RIRD-Real \textit{Meeting Room} scene. We report the per-frame runtime, FPS, and number of Gaussians for each rendering stage.
}
\Description{
A table reporting the rendering speed of the proposed system on the RIRD-Real Meeting Room scene. The rows include TSDF raycasting, base Gaussian rendering, reflection Gaussian rendering, and total rendering. The table reports runtime per frame, FPS, and the number of Gaussians used in each Gaussian rendering stage.
}
\footnotesize
\resizebox{\columnwidth}{!}{
\begin{tabular}{lccc}
\toprule
Rendering Stage & Time / Frame (ms) & FPS & Num. of Gaussians \\
\hline
TSDF Raycasting              & 0.72 & 1389.83 & -- \\
Base Gaussian Rendering      & 1.05 & 950.00  & 0.30M \\
Reflection Gaussian Rendering & 1.39 & 716.19  & 0.34M \\
Total                        & 3.22 & 310.00  & 0.64M \\
\bottomrule
\end{tabular}
}
\label{tab:render_speed}
\end{table}

\subsubsection{Reflective plane identification}

Fig.~\ref{fig:plane_detect} visualizes representative results of reflective plane identification. From left to right, we show the input RGB-D frames, plane segmentation maps, semantic segmentation maps, TSDF color variance maps, and predicted reflection masks. The examples show that combining geometric, semantic, and temporal cues enables the system to localize reflective regions on planar surfaces.

\subsection{Ablation Studies}

\begin{figure}[t]
    \centering
    \includegraphics[width= \columnwidth]{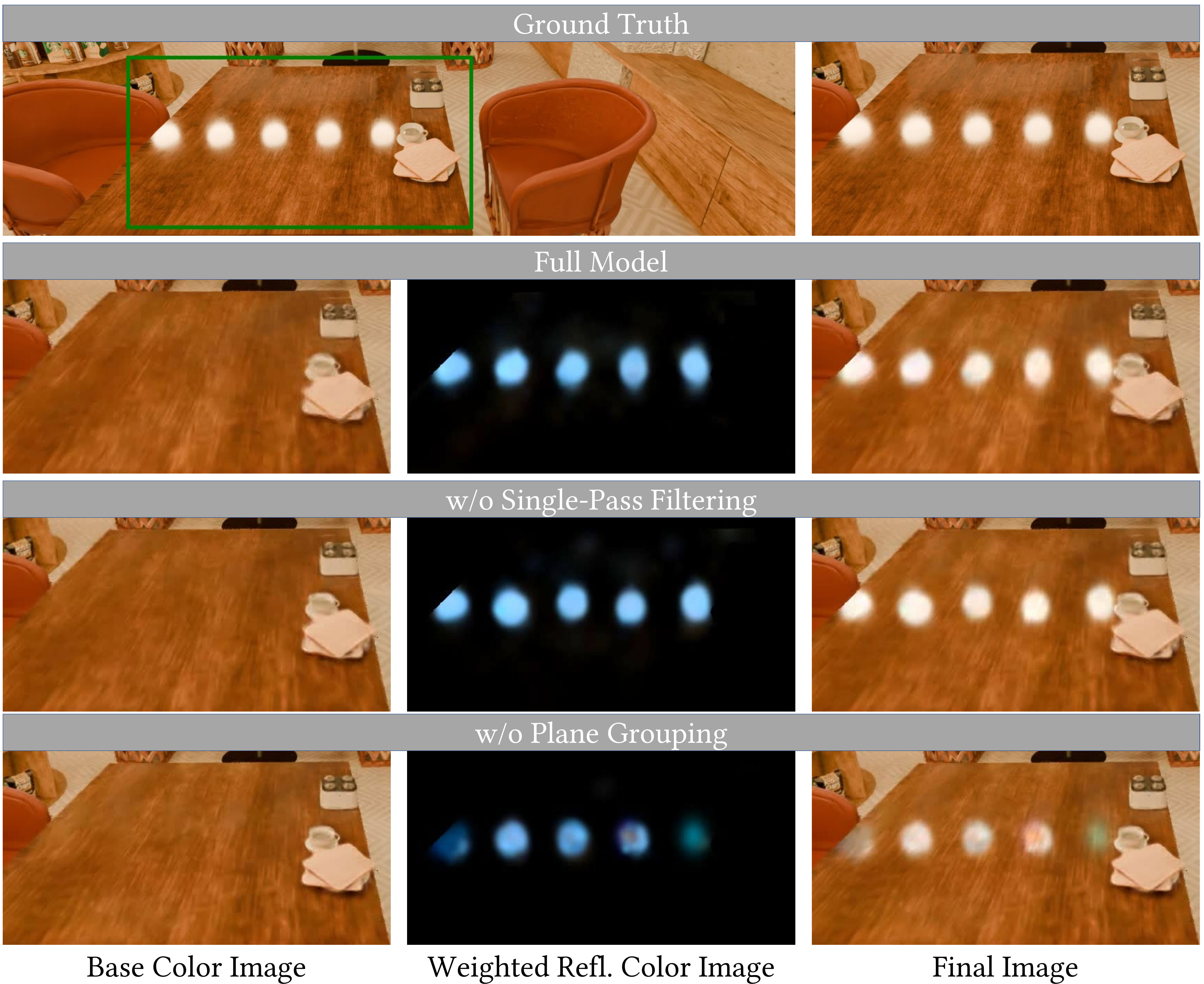}
    \caption{
    Ablation study of plane-conditioned reflection rendering on the \textit{Room 2} scene from RIRD-Syn. Without plane grouping, reflections become dim and blurry. Plane-associated reflection groups produce more accurate reflected content, while our single-pass plane-filtered rendering achieves comparable quality to multi-pass masked compositing with higher efficiency.
    }
    \Description{ A qualitative ablation figure comparing reflection rendering results on a synthetic indoor scene. The figure shows results from variants without plane grouping, without single-pass filtering, and the full model. The variant without plane grouping produces weaker and blurrier reflections, while the plane-conditioned variants produce clearer and more accurate reflected content. The full model preserves similar reflection quality to the multi-pass variant while using a more efficient single-pass rendering strategy. }                     
    \label{fig:ablation_refl_render}
\end{figure}

\subsubsection{Reflection grouping and rendering}
To evaluate the effectiveness of our reflection representation, we ablate two key designs: plane-associated reflection grouping, which decomposes reflections according to their supporting reflective planes, and single-pass plane filtering, which restricts each reflection group to its corresponding planar region during rendering. We conduct the ablation on the \textit{Room 2} scene from RIRD-Syn. We compare three variants: \textit{w/o plane grouping}, which represents the whole scene with a single set of reflection Gaussians and renders them as in vanilla 3DGS; \textit{w/o single-pass filtering}, which keeps plane-associated reflection groups but renders them separately and composites the results using plane masks; and the full model, which renders all reflection groups in a single pass with per-pixel plane filtering. As shown in Fig.~\ref{fig:ablation_refl_render}, \textit{w/o plane grouping} produces dimmer and blurrier reflections. Both \textit{w/o single-pass filtering} and the full model produce more accurate reflection colors using plane-associated groups, but the former requires additional rendering passes and is much slower, as shown in Table~\ref{table:ablation_refl_render}. These results show that reflection grouping improves reflection fidelity, while single-pass plane filtering preserves this quality with substantially higher rendering efficiency.

\begin{table}[t]
\centering
\caption{
Ablation study of reflection grouping and rendering on RIRD-Syn \textit{Room 2}. We report base rendering time, reflection rendering time, total rendering FPS, and PSNR. The results show that plane grouping improves reflection quality, while our single-pass filtering maintains high rendering efficiency.
}
\Description{
A table reporting the ablation study of reflection grouping and rendering. The compared variants include removing plane grouping, removing single-pass filtering, and the full model. The table reports base rendering time, reflection rendering time, rendering FPS, and PSNR. The full model achieves the best PSNR and rendering FPS while keeping reflection rendering efficient.
}
\footnotesize
\begin{tabular}{l|cccc}
\toprule
Ablation & \makecell{Base Render\\(ms)$\downarrow$} & \makecell{Refl. Render\\(ms)$\downarrow$} & \makecell{Render\\FPS$\uparrow$} & PSNR$\uparrow$ \\ 
\hline
w/o Plane Grouping        & 3.37      & 0.33      & 265.4      & 26.55 \\ 
w/o Single-Pass Filtering & \uli{3.32} & 1.39       & 209.2      & 26.90 \\
Full Model                & 3.36      & \uli{0.29} & \uli{268.8} & \uli{27.09} \\ 
\bottomrule
\end{tabular}
\label{table:ablation_refl_render}
\end{table}

\subsubsection{Reflection-aware tracking}
\begin{figure}[t]
    \centering
    \includegraphics[width=\columnwidth]{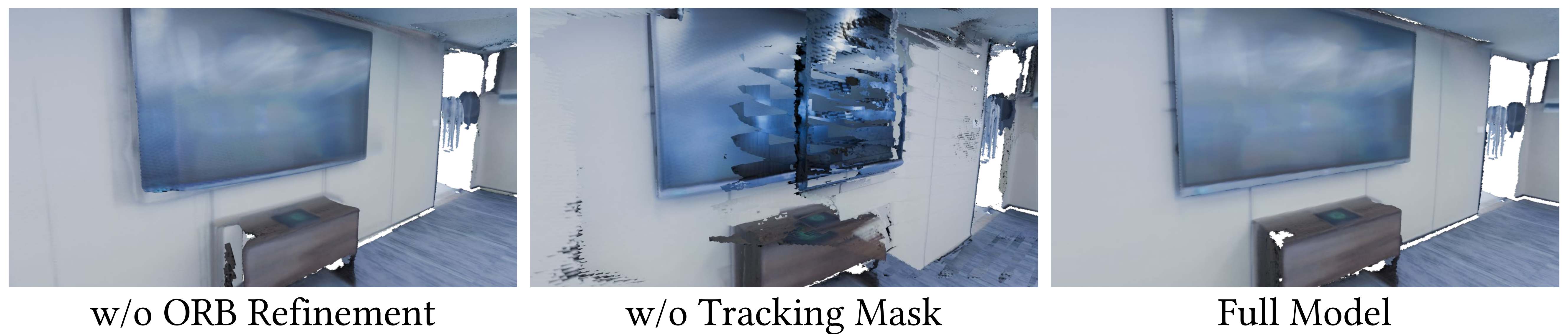}
    \caption{
    Ablation study of reflection-aware tracking. We compare three variants: \textit{w/o ORB refinement}, which uses ICP only; \textit{w/o tracking mask}, which applies ORB refinement without masking reflection-dominated regions; and the \textit{Full model}, which uses masked ORB refinement. The results show that unmasked ORB refinement can introduce drift in reflective regions, while the full model produces cleaner reconstruction.
    }
    \Description{
    A qualitative ablation figure for reflection-aware tracking. The figure compares reconstructed meshes under three variants: ICP-only tracking without ORB refinement, ORB refinement without reflection masking, and the full model with reflection-aware masked ORB refinement. The variant without reflection masking shows visible drift in reflective regions, while the full model produces a cleaner and more stable reconstruction.
    }
    \label{fig:ablation_track}
\end{figure}

We ablate the proposed reflection-aware tracking strategy on the \textit{Room 1} scene from RIRD-Syn and the \textit{Meeting Room} scene from RIRD-Real. We compare three variants: \textit{w/o ORB refinement}, which uses ICP only; \textit{w/o tracking mask}, which applies ORB-based pose refinement without masking reflection-dominated regions; and the \textit{full model}, which uses masked ORB-based pose refinement. On RIRD-Syn, adding ORB-based pose refinement without masking increases the ATE from 0.56 to 1.21, indicating that unreliable features from reflection-dominated regions can degrade pose estimation. With reflection-aware masking, the ATE is reduced to 0.31, showing that suppressing reflective regions allows feature-based pose refinement to improve tracking robustness. Since the synthetic data are relatively idealized and show less visible mesh artifacts, we further evaluate the reconstructed meshes on the real \textit{Meeting Room} scene in Fig.~\ref{fig:ablation_track}. The full model produces cleaner reconstruction, whereas the variant without reflection masking exhibits visible drift.

\begin{table}[h]
\caption{
Quantitative ablation study of reflection-aware tracking on RIRD-Syn \textit{Room 1}. We report ATE for tracking accuracy and geometry metrics including accuracy, completion, and their corresponding ratios.
}
\Description{
A quantitative ablation table evaluating reflection-aware tracking. The compared variants include removing ORB refinement, removing the reflection mask, and the full model. The table reports ATE, geometry accuracy, accuracy ratio, completion, and completion ratio. The full model achieves the best tracking accuracy and geometry reconstruction quality.
}
\footnotesize
\begin{tabular}{c|c|cccc}
\toprule[1pt]
Ablation & ATE$\downarrow$ & Acc.$\downarrow$ & Acc. Ratio$\uparrow$ & Com.$\downarrow$ & Com. Ratio$\uparrow$  \\ \hline
w/o ORB Refinement         & 0.56    & 0.95      & 97.54        & 0.96         &  98.64             \\
w/o Tracking Mask        & 1.21      & 3.19      & 86.72        & 0.93         & 92.52      \\
Full Model & \uli{0.31} & \uli{0.93} & \uli{98.86}   & \uli{0.81} & \uli{99.82}     \\
\bottomrule[1pt]
\end{tabular}
\label{table:ablation_track}
\end{table}

\begin{figure}[t]
    \centering
    \includegraphics[width=\columnwidth]{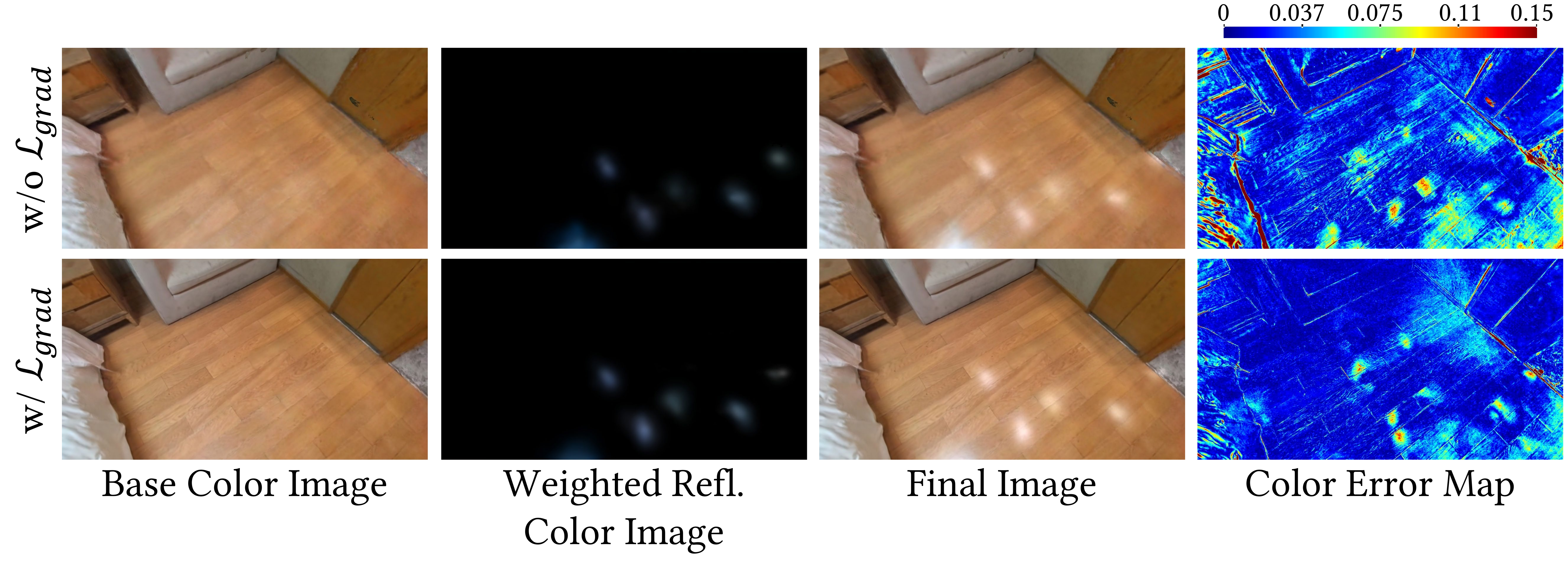}
    \caption{
    Ablation study of the reflection color regularization term on RIRD-Real \textit{Living Room 0}. We compare the results without and with $\mathcal{L}_{grad}$. From left to right, we show the base color image, weighted reflection color image, final composited image, and color error map. With $\mathcal{L}_{grad}$, the base and reflection components are better constrained during optimization, resulting in lower color errors and improved final rendering quality.
    }
    \Description{
    A qualitative ablation figure comparing the method without and with the reflection color regularization term on the RIRD-Real Living Room 0 scene. The top row shows results without the regularization term, and the bottom row shows results with the regularization term. The columns show the base color image, weighted reflection color image, final composited image, and color error map. Without the regularization term, the base and reflection components exhibit stronger interference and the color error map shows larger errors. With the regularization term, the color error is reduced and the final rendering quality is improved.
    }
    \label{fig:ablation_grad_loss}
\end{figure}

\subsubsection{Effect of Regularization on Reflection Color}
We evaluate the effect of the reflection color regularization term $\mathcal{L}_{grad}$ on the RIRD-Real \textit{Living Room 0} scene by comparing our full model with a variant without this term. As shown in Table~\ref{table:ablation_grad_loss}, removing $\mathcal{L}_{grad}$ degrades rendering quality. Fig.~\ref{fig:ablation_grad_loss} further shows that, without this regularization on the reflection color, the base color becomes blurrier, and the final composited image loses high-frequency details. This is because the base and reflection components are less constrained during joint optimization and tend to interfere with each other. The color error visualization further confirms that $\mathcal{L}_{grad}$ reduces such interference and improves the final rendering quality.

\begin{table}[t]
\caption{
Ablation study of the reflection color regularization term on RIRD-Real \textit{Living Room 0}. We compare the rendering quality with and without $\mathcal{L}_{grad}$.
}
\Description{
A table comparing the effect of the reflection color regularization term on the Living Room 0 scene from RIRD-Real. Without the regularization term, the method achieves 28.36 PSNR, 0.888 SSIM, and 0.231 LPIPS. With the regularization term, the method improves to 30.38 PSNR, 0.919 SSIM, and 0.224 LPIPS.
}
\footnotesize
\begin{tabular}{c|ccc}
\toprule
Ablation              & PSNR$\uparrow$  & SSIM$\uparrow$  & LPIPS$\downarrow$ \\
\hline
w/o $\mathcal{L}_{grad}$ & 28.36 & 0.888 & 0.231 \\
w/  $\mathcal{L}_{grad}$  & \uli{30.38} & \uli{0.919} & \uli{0.224} \\
\bottomrule
\end{tabular}
\label{table:ablation_grad_loss}
\end{table}

\subsubsection{Effect of Temporal Color Variance Threshold}
We analyze the influence of the temporal color variance threshold $\delta_v$ used for reflective plane verification on the \textit{Room 2} scene from RIRD-Syn. By varying $\delta_v$, we evaluate its impact on novel-view rendering quality. As shown in Table~\ref{table:ablation_colorvar}, our method achieves stable performance across a moderate range of threshold values, demonstrating that the proposed temporal reflection verification is robust to threshold selection. When $\delta_v$ becomes too large, reflective regions may be missed, leading to degraded rendering quality.


\begin{table}[t]
\caption{
Ablation study of the temporal color variance threshold $\delta_v$ on RIRD-Syn \textit{Room 2}. We report PSNR, SSIM, and LPIPS under different threshold values to evaluate the sensitivity of reflective-region verification.
}
\Description{
A quantitative ablation table evaluating the temporal color variance threshold used for reflective-region verification. The rows correspond to different values of delta v, and the columns report PSNR, SSIM, and LPIPS. The results show that the method remains stable across a moderate range of threshold values.
}
\footnotesize
\begin{tabular}{l|ccc}
\toprule[1pt]
Threshold          & PSNR$\uparrow$ & SSIM$\uparrow$ & LPIPS$\downarrow$  \\ \hline
$\delta_v{=}0.002$   & 26.23     & 0.872   & 0.225              \\
$\delta_v$=0.005   & 26.37     & 0.881   & 0.214              \\
$\delta_v$=0.01    & \uli{26.47}     & \uli{0.883}   & 0.211                   \\
$\delta_v$=0.03    & 26.27     & 0.879   & \uli{0.209}                   \\
$\delta_v$=0.05    & 25.65     & 0.873   & 0.253                \\
$\delta_v$=0.1     & 24.06     & 0.864   & 0.243      \\
\bottomrule[1pt]
\end{tabular}
\label{table:ablation_colorvar}
\end{table}

\section{Conclusion}
We present the first real-time RGB-D Gaussian SLAM system designed for indoor scenes with planar reflections. We introduce a reflection-aware TSDF-Gaussian hybrid representation that explicitly decomposes the scene into a base component and plane-associated reflection components, enabling more faithful modeling of reflective appearance during online reconstruction. To render this representation efficiently, we develop a three-pass pipeline that combines TSDF raycasting, base Gaussian rendering, and plane-conditioned reflection Gaussian rasterization. For online reconstruction, our system integrates reflection-aware tracking, reflective plane identification from geometric, semantic, and temporal cues, augmented TSDF fusion, and online optimization of both base and reflection Gaussians. Experiments on self-captured reflective scenes, a synthetic reflective dataset, and public RGB-D benchmarks demonstrate that our method achieves superior reconstruction quality, more robust tracking, and higher-quality novel-view rendering in indoor environments with reflections while maintaining real-time performance.

Our method currently focuses on planar reflections and therefore cannot explicitly model reflections on curved surfaces, whose geometry and view-dependent appearance are substantially more complex. In addition, transparent objects without reliable depth measurements, such as glass, remain challenging for our RGB-D reconstruction pipeline and may lead to incomplete geometry or unstable appearance modeling. Extending reflection-aware online reconstruction to curved reflective surfaces, transparent materials, and more general non-Lambertian effects would be valuable directions for future work.

\bibliographystyle{ACM-Reference-Format}
\bibliography{texts/bib}



\end{document}


\settopmatter{printacmref=false, printccs=false}
\setcopyright{none}
\renewcommand\footnotetextcopyrightpermission[1]{}

\title{Supplementary Material for RRG-SLAM: Real-time Reflection-aware Gaussian SLAM for Indoor
Scenes}

\maketitle








\appendix

\section{Dataset details}
RIRD-Real is a real-world indoor RGB-D dataset collected by ourselves in reflective environments containing various reflective materials and objects. All frames have a resolution of 1280×720. The dataset includes diverse reflective surfaces such as glossy floors, tables, and TVs commonly found in indoor scenes. Table~\ref{dataset:rird_real} summarizes the statistics of RIRD-Real.
\begin{table}[h]
\caption{
Statistics of the RIRD-Real dataset. LR0 and LR1 denote the two living-room scenes. We report the number of frames, trajectory length, and scanned area for each scene.
}
\Description{
A table summarizing the RIRD-Real dataset. The columns correspond to six real indoor scenes, including two living rooms, a restroom, a meeting room, a private room, and a kitchen. The rows report the number of frames, trajectory length in meters, and scanned area in square meters for each scene.
}
\footnotesize
\begin{tabular}{ccccccc}
\toprule
Statistic             & LR0 & LR1 & Rest & Meeting & Private   & Kitchen \\ 
\hline
Frames                & 1570          & 2080          & 3470      & 3699         & 1649  & 2670    \\
Traj. Len. (m) & 15.3          & 20.1          & 33.7    & 59.1         & 20.7  & 26.4    \\
Area (m$^2$)        & 19.2          & 15.2         & 41.7     & 102.2       & 21.6 & 36.7   \\
\bottomrule
\end{tabular}

\label{dataset:rird_real}
\end{table}

RIRD-Syn is a synthetic RGB-D dataset rendered with Blender at a resolution of 1280×720, enabling quantitative evaluation of novel-view synthesis and tracking. Table~\ref{dataset:rird_syn} summarizes the statistics of RIRD-Syn. We also select two reflective scenes from ScanNet++~\cite{yeshwanth2023scannet++} with strong planar reflections, denoted as a5859cfd40 and 8e6ff28354, at a resolution of 876×584.
\begin{table}[h]
\caption{
Statistics of the RIRD-Syn dataset. We report the number of frames, trajectory length, and scanned area for each synthetic scene.
}
\Description{
A table summarizing the RIRD-Syn dataset. The columns correspond to three synthetic indoor scenes, Room 0, Room 1, and Room 2. The rows report the number of frames, trajectory length in meters, and scanned area in square meters for each scene.
}
\footnotesize
\begin{tabular}{cccc}
\toprule
Statistic             & Room 0 & Room 1 & Room 2  \\ 
\hline
Frames                & 1681          & 2337          & 2206         \\
Traj. Len. (m) & 41.9          & 58.5        & 50.6       \\
Area (m$^2$)       & 43.8          & 43.25         & 97.8      \\
\bottomrule
\end{tabular}
\label{dataset:rird_syn}
\end{table}

\section{Implementation Details}
Across all datasets, reflective plane identification is performed every $N_p=10$ frames, and Gaussian optimization is executed every $N_o=10$ frames. During optimization, we use $N_{local}=2$ recent frames together with $N_{global}=7$ randomly sampled keyframes, and perform 30 optimization iterations for each optimization step. For TSDF fusion, we use a voxel size of 1 cm for the 102.2 m$^2$ \textit{Meeting Room} scene and 0.5 cm for all other scenes.

For Gaussian densification, we use different thresholds for base and reflection Gaussians. For base Gaussians, we set the color error threshold $\delta_b$ to 0.05 and the Gaussian weight threshold $\delta_{w,b}$ to 3. For reflection Gaussians, we set the color error threshold $\delta_r$ to 0.05 and the reflection Gaussian weight threshold $\delta_{w,r}$ to 0.5.

For Gaussian optimization, we use different learning rates for base and reflection Gaussians. For base Gaussians, we set the learning rates $lr_{position}^b=0.00016$, $lr_{scale}^b=0.005$, $lr_{opacity}^b=0.05$ $lr_{rotation}^b=0.001$, $lr_{SH_0}^b=0.0025$, and $lr_{SH_{rest}}^b=0.0005$. For reflection Gaussians, we set the learning rates $lr_{position}^r=0.00016$, $lr_{scale}^r=0.005$, $lr_{oapcity}^r=0.001$, $lr_{rotation}^r=0.001$, $lr_{SH_0}^r=0.0025$, and $lr_{SH_{rest}}^r=0.0005$.

For Gaussian pruning, we remove Gaussians according to their opacity and scale. Specifically, Gaussians are pruned when their opacity is lower than the opacity threshold, or when their scale is smaller than the minimum scale threshold or larger than the maximum scale threshold. For base Gaussians, the opacity threshold, minimum scale threshold, and maximum scale threshold are set to 0.005, 0.003, and 0.1, respectively. For reflection Gaussians, the corresponding thresholds are set to 0.005, 0.005, and 0.2.

\section{TSDF Color Variance Update}

To identify temporally unstable reflective regions, we maintain a color inconsistency statistic for each TSDF voxel. In addition to the fused RGB color, each voxel stores a running mean $\mu$ and a second central moment $M_2$ of the observed color intensity.

Given a new RGB observation $\mathbf{c}_t=(r_t,g_t,b_t)$ at frame $t$, we first normalize it to $[0,1]$ and convert it into a scalar intensity:
\begin{equation}
x_t=\frac{r_t+g_t+b_t}{3}.
\end{equation}

The fused RGB color is updated using standard running averaging:
\begin{equation}
\bar{\mathbf{c}}_{n+1}
=
\frac{
n\bar{\mathbf{c}}_n+\mathbf{c}_t
}{
n+1
}.
\end{equation}

Here, $t$ denotes the current frame index, and $n$ denotes the number of accumulated color observations for the corresponding voxel before incorporating $\mathbf{c}_t$.

The color intensity statistics are then updated online using a Welford-style~\cite{welford1962note} update:
\begin{equation}
\delta_t=x_t-\mu_n,
\end{equation}
\begin{equation}
\mu_{n+1}
=
\mu_n+\frac{\delta_t}{n+1},
\end{equation}
\begin{equation}
M_{2,n+1}
=
M_{2,n}+\delta_t(x_t-\mu_{n+1}).
\end{equation}

After incorporating the new observation, the per-voxel color variance is computed as
\begin{equation}
\sigma_c^2
=
\frac{M_{2,n+1}}{n+1},
\end{equation}
where $\sigma_c^2$ denotes the temporal variance of the observed color intensity for the voxel.

During TSDF raycasting, the variance values from neighboring voxels are trilinearly interpolated to produce a dense color variance map:
\begin{equation}
\mathbf{C}_{\mathrm{var}}(\mathbf{x})
=
\frac{
\sum_i w_i\sigma_{c,i}^2
}{
\sum_i w_i+\epsilon
},
\end{equation}
where $\mathbf{x}$ denotes the raycast surface point, $w_i$ denotes the trilinear interpolation weight of the $i$-th neighboring voxel, and $\epsilon$ is a small constant for numerical stability.

Pixels whose raycast variance exceeds a predefined threshold $\delta_v$ are treated as temporally unstable regions and are further used for reflective-plane verification and reflection Gaussian initialization.


\begin{table}[t]
\caption{
Per-scene tracking accuracy comparison on RIRD-Syn. We report ATE for each synthetic reflective scene and the average across all three scenes.
}
\Description{
A quantitative table comparing tracking accuracy on the RIRD-Syn dataset. The rows list different SLAM methods, and the columns report ATE on Room 0, Room 1, and Room 2, together with the average performance. The proposed method achieves the lowest ATE on all three synthetic reflective scenes and the best average tracking accuracy.
}
\footnotesize
\begin{tabular}{ccccc}
\toprule
Method       & Room 0                     & Room 1                     & Room 2                     & Average                    \\
\hline
ORB-SLAM2    & 63.45                      & 1.35                       & 1.98                      & 22.26                      \\ 
BundleFusion & 22.40                      & 0.92                       & 2.23                       & 8.52                       \\
SplaTAM      & 103.32                     & 3.00                       & 2.13                       & 36.15                      \\
RTG-SLAM     & 30.24                      & 0.43                  & 1.46 & 10.71                      \\
GauS-SLAM    & 116.76                     & 6.64                       & 2.03                       & 41.81                      \\
GS-ICP SLAM  & 25.00                      & \cs{0.29} & 24.56                      & 16.62                      \\
GPS-SLAM     & \cs{0.63} & 0.69                     & \cs{0.74}                       & \cs{0.69} \\
Ours         & \cf{0.23} & \cf{0.28} & \cf{0.42} & \cf{0.31} \\
\bottomrule
\end{tabular}
\label{table:ate_rird_syn}
\end{table}

\begin{table}[t]
\caption{
Per-sequence tracking accuracy comparison on TUM-RGBD. We report ATE for each evaluated sequence and the average, where lower values indicate better tracking accuracy.
}
\Description{
A quantitative table comparing tracking accuracy on the TUM-RGBD dataset. The rows list different SLAM methods, and the columns report ATE on fr1\_desk, fr2\_xyz, and fr3\_office, together with the average performance. The proposed method achieves the lowest average ATE among the compared methods.
}
\footnotesize
\begin{tabular}{c|ccc|c}
\toprule
Method       & fr1\_desk & fr2\_xyz & fr3\_office & Average \\ \hline
ORB-SLAM2    & 1.62 & \cs{0.45}     & 1.75        & 1.27    \\
BundleFusion & 1.65      & 1.13      & 1.81        & 1.53    \\
SplaTAM      & 1.62      & 1.30      & 2.53        & 1.81    \\
RTG-SLAM     & \cs{1.51}      & 0.58      & \cs{1.32}        & \cs{1.13}    \\
GauS-SLAM    & \cf{1.48}      & 1.29      & 1.50        & 1.42    \\
GS-ICP SLAM  & 2.96      & 1.77      & 2.26        & 2.33    \\
GPS-SLAM     & 3.5       & 2.60      & 2.15        & 2.75    \\
Ours         & 1.61 & \cf{0.38} & \cf{1.17}        & \cf{1.06}    \\ \bottomrule
\end{tabular}
\label{table:ate_tum}
\end{table}

\begin{table}[t]
\caption{
Per-scene tracking accuracy comparison on Replica. We report ATE for each evaluated room and office scene, together with the average across all scenes.
}
\Description{
A quantitative table comparing tracking accuracy on the Replica dataset. The rows list different SLAM methods, and the columns report ATE on three room scenes and five office scenes, together with the average performance. The proposed method achieves competitive tracking accuracy on this general indoor benchmark.
}
\resizebox{0.95\columnwidth}{!}{
\begin{tabular}{c|cccccccc|c}
\toprule
Method       & Rm 0      & Rm 1      & Rm 2      & Off 0     & Off 1     & Off 2     & Off 3     & Off 4     & Avg.      \\
\hline
ORB-SLAM2    & 0.34      & 0.38      & 0.38      & 0.33      & 0.32      & 0.41      & 0.34      & 0.37      & 0.36      \\
BundleFusion & 0.41      & 0.40      & 0.43      & 0.43      & 0.37      & 0.42      & 0.46      & 0.44      & 0.42      \\
SplaTAM      & 0.38      & 0.23      & 0.28      & 0.48      & 0.32      & 0.38      & 0.43      & 0.64      & 0.39      \\
RTG-SLAM     & 0.20      & 0.19      & \cs{0.12} & \cs{0.16} & 0.13      & 0.22      & 0.25      & 0.25      & 0.19      \\
GauS-SLAM    & \cf{0.06} & \cf{0.08} & \cf{0.07} & \cf{0.06} & \cf{0.04} & \cf{0.09} & \cf{0.07} & \cf{0.06} & \cf{0.07} \\
GS-ICP SLAM  & 0.16      & 0.17      & 0.12      & 0.19      & \cs{0.13} & \cs{0.17} & \cs{0.18} & 0.23      & 0.18      \\
GPS-SLAM     & 0.18      & 0.19      & 0.17      & 0.17      & 0.14      & 0.25      & 0.20      & 0.21      & 0.19      \\
Ours         & \cs{0.16} & \cs{0.16} & 0.17      & 0.16      & 0.13      & 0.18      & 0.19      & \cs{0.18} & \cs{0.17} \\
\bottomrule
\end{tabular}
}
\label{table:ate_replica}
\end{table}

\section{Global Plane Maintenance}
During online reconstruction, each frame first obtains local planar segments from CAPE~\cite{cape} together with their plane parameters. Since these local plane labels are only valid within the current frame, we incrementally register them into a globally maintained plane table to obtain temporally consistent plane IDs across frames.

For each detected local plane, we first apply a geometric filter against all existing global planes: a match requires the normal angle difference to be within $10^\circ$ and the point-to-plane distance to be below $0.1$\,m. However, geometric consistency alone cannot distinguish distinct coplanar instances that share the same infinite plane (e.g., two table segments separated by a gap). We therefore maintain a spatial footprint for each global plane to record its occupied region on the plane surface.

To build the footprint, we construct a 2D orthonormal basis on the plane and discretize the surface into a regular grid with cell size $0.05$\,m. Each 3D point belonging to the plane is projected onto this grid, and the occupied cells along with their axis-aligned bounding box are stored. When associating a local plane, we project its points into each candidate global plane's coordinate frame and compute three metrics: the cell overlap ratio (with 1-cell dilation), the bounding box IoU, and the bounding box gap. A match is established if the overlap exceeds $0.05$, the IoU exceeds $0.02$, or the gap is below $0.10$m for a recently observed instance. Among all matching candidates, we select the one with the highest overlap to prevent a noisy segment from bridging two distinct instances. If no candidate matches, a new global instance is created.

Over time, instances initially created separately may be recognized as the same physical plane. We use a disjoint-set union structure to merge them: the lower-index instance absorbs the higher-index one by merging their plane parameters, unifying their footprints and bounding boxes, and retiring the absorbed instance.

Finally, we maintain a plane ID table that remaps all plane indices to the canonical global IDs after merging. During TSDF fusion and raycasting, all plane indices are converted through this table to ensure consistent plane association. The resulting global plane map is used for reflection-region association, reflection Gaussian initialization, and plane-conditioned reflection rendering.

\begin{figure*}[h]
    \centering
    \includegraphics[width=0.95\textwidth,height=0.95\textheight,keepaspectratio]{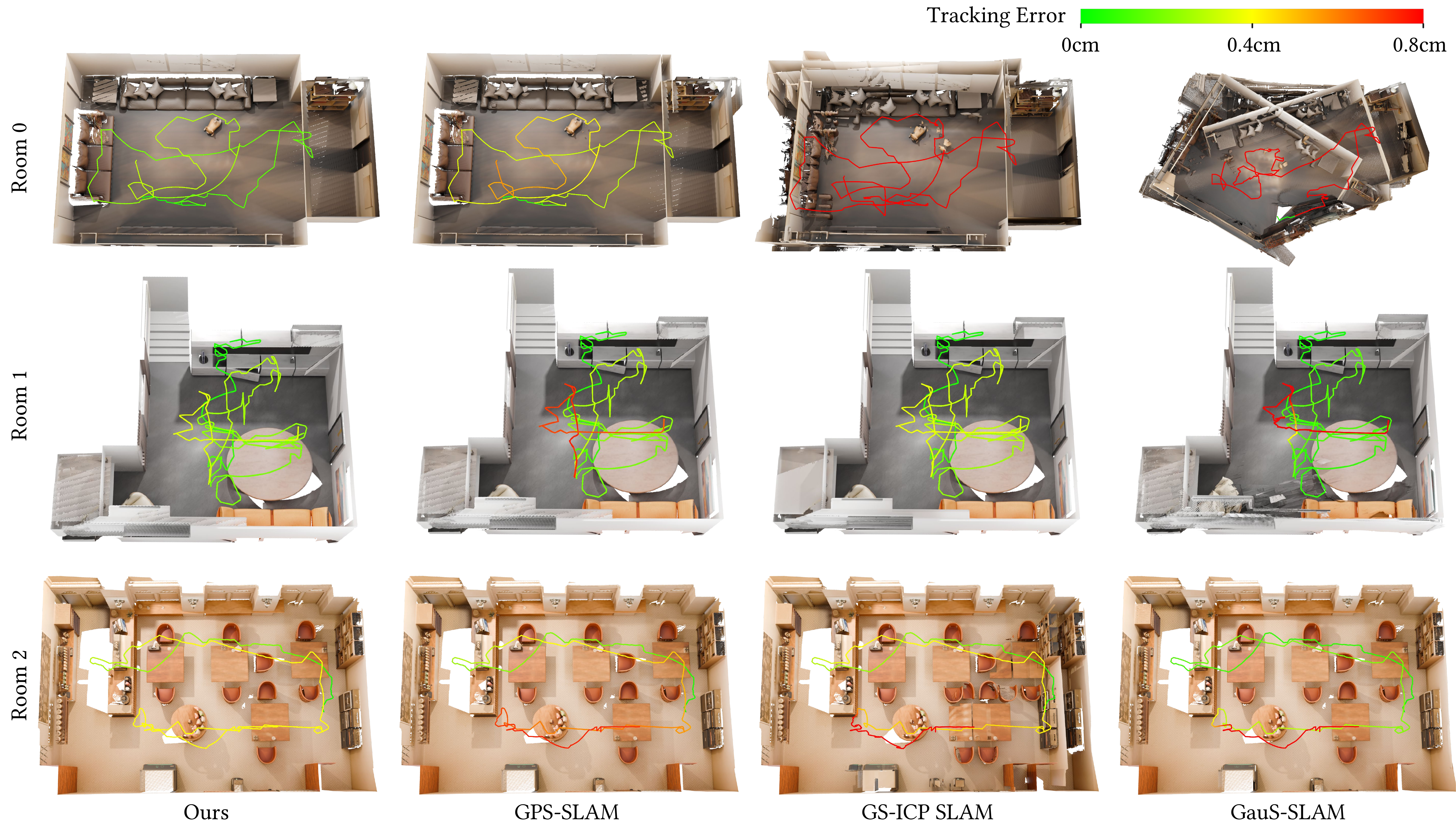}
    \caption{
    Tracking-error and mesh visualization on RIRD-Syn. We compare our method with GPS-SLAM, GS-ICP SLAM, and GauS-SLAM on three synthetic reflective scenes. The camera trajectories are colored by frame-wise tracking error, where green indicates lower error and red indicates larger error. Our method maintains stable trajectories and produces more coherent mesh reconstruction in reflective scenes.
    }
    \Description{
   A qualitative comparison figure showing reconstructed meshes and tracking-error-colored camera trajectories on three RIRD-Syn scenes: Room 0, Room 1, and Room 2. The columns compare Ours, GPS-SLAM, GS-ICP SLAM, and GauS-SLAM. A color bar indicates tracking error from green to red, corresponding to low to high error. The proposed method shows more stable trajectories and cleaner reconstructed meshes, while some baseline methods exhibit larger tracking errors and visible reconstruction distortions.
    }
    \label{fig:tracking_error_vis}
\end{figure*}

\section{Grounded SAM2 Settings}
We use the official implementation of Grounded SAM2~\cite{groundingdino,sam2} for semantic reflective-region detection, with all parameters following the default settings provided by the official codebase. Given predefined text prompts, GroundingDINO~\cite{groundingdino} first predicts object bounding boxes, which are then passed to SAM2~\cite{sam2} to obtain the final segmentation masks.

The text prompts used in our implementation are: “floor”, “table”, “vending machine”, “closet”, “wall art”, “tv”, and “whiteboard”. These semantic masks are used only to generate reflective-plane candidates before temporal reflection verification.

\section{More Results}

\subsection{Per-scene Tracking Accuracy}
We report detailed per-scene results for the same baselines used in the main paper, including ORB-SLAM2~\cite{orbslam2}, BundleFusion~\cite{bundlefusion}, SplaTAM~\cite{splatam}, RTG-SLAM~\cite{rtg-slam}, GauS-SLAM~\cite{gaus-slam}, GS-ICP SLAM~\cite{gs-icp}, and GPS-SLAM~\cite{gps-slam}.
We provide detailed per-scene tracking accuracy results on RIRD-Syn, TUM-RGBD~\cite{tum}, and Replica~\cite{straub2019replica}. Tables~\ref{table:ate_rird_syn}, \ref{table:ate_tum}, and \ref{table:ate_replica} report the absolute trajectory error (ATE) for each evaluated sequence. These results complement the averaged tracking results in the main paper and provide a more detailed comparison of tracking robustness across reflective and general indoor scenes.

\begin{table}[t]
\centering
\caption{
Per-scene novel-view rendering quality comparison on the selected reflective ScanNet++ scenes. We report PSNR, SSIM, and LPIPS for each scene and the average across the two scenes.
}
\Description{
A quantitative table comparing novel-view rendering quality on two selected reflective ScanNet++ scenes. The rows list different SLAM methods, and each method is evaluated using PSNR, SSIM, and LPIPS on scenes a5859cfd40 and 8e6ff28354, together with the average performance. The proposed method achieves the best average rendering quality across the reported metrics.
}
\footnotesize
\begin{tabular}{c|c|cc|c}
\toprule
Method      & Metric  & a5859cfd40                  & 8e6ff28354                  & Average                     \\
            \hline
\multirow{3}{*}{SplaTAM}     & PSNR$\uparrow$   & 21.08                       & 26.63                       & 23.85                       \\
            & SSIM$\uparrow$   & 0.781                       & 0.887                       & 0.834                       \\
            & LPIPS$\downarrow$  & 0.293                       & 0.190                       & 0.241                       \\ \hline
\multirow{3}{*}{RTG-SLAM}    & PSNR$\uparrow$   & 18.67                       & 26.07                       & 22.37                       \\
            & SSIM$\uparrow$   & 0.705                       & 0.862                       & 0.783                       \\
            & LPIPS$\downarrow$  & 0.372                       & 0.226                       & 0.299                       \\ \hline
\multirow{3}{*}{GauS-SLAM}   & PSNR$\uparrow$   & \cs{22.75} & \cs{27.59} & \cs{25.17} \\
            & SSIM$\uparrow$   & 0.745                       & 0.856                       & 0.800                       \\
            & LPIPS$\downarrow$  & 0.332                       & 0.243                       & 0.287                       \\ \hline
\multirow{3}{*}{GS-ICP SLAM} & PSNR$\uparrow$   & 21.96                       & 27.50                       & 24.73                       \\
            & SSIM$\uparrow$   & \cs{0.821} & \cs{0.913} & \cs{0.867} \\
            & LPIPS$\downarrow$  & \cs{0.238} & \cs{0.149} & \cs{0.194} \\ \hline
\multirow{3}{*}{GPS-SLAM}    & PSNR$\uparrow$   & 22.18                       & 26.99                       & 24.59                       \\
            & SSIM$\uparrow$   & 0.781                       & 0.890                       & 0.836                       \\
            & LPIPS$\downarrow$  & 0.317                       & 0.193                       & 0.255                       \\ \hline
\multirow{3}{*}{Ours}        & PSNR$\uparrow$   & \cf{23.51} & \cf{28.50} & \cf{26.01} \\
            & SSIM$\uparrow$   & \cf{0.823} & \cf{0.916} & \cf{0.869} \\
            & LPIPS$\downarrow$  & \cf{0.229} & \cf{0.141} & \cf{0.185} \\
\bottomrule
\end{tabular}
\label{table:rq_scannetpp}
\end{table}

\begin{table}[t]
\caption{
Per-scene novel-view rendering quality comparison on RIRD-Syn. We report PSNR, SSIM, and LPIPS for each synthetic reflective scene and the average across all three scenes.
}
\Description{
A quantitative table comparing novel-view rendering quality on the RIRD-Syn dataset. The rows list different SLAM methods, and each method is evaluated using PSNR, SSIM, and LPIPS on Room 0, Room 1, and Room 2, together with the average performance. The proposed method achieves the best average rendering quality across the reported metrics.
}
\footnotesize
\begin{tabular}{c|c|ccc|c}
\toprule
Method                       & Metric            & Room 0     & Room 1     & Room 2     & Average    \\
\hline
\multirow{3}{*}{SplaTAM}     & PSNR$\uparrow$    & 15.43      & 23.45      & 17.08      & 18.66      \\
                             & SSIM$\uparrow$    & 0.592      & 0.764      & 0.547      & 0.634      \\
                             & LPIPS$\downarrow$ & 0.474      & 0.341      & 0.453      & 0.422      \\ 
                             \hline
\multirow{3}{*}{RTG-SLAM}    & PSNR$\uparrow$    & 14.74      & 22.97      & 15.74      & 17.81      \\
                             & SSIM$\uparrow$    & 0.605      & 0.813      & 0.533      & 0.650      \\
                             & LPIPS$\downarrow$ & 0.496      & 0.332      & 0.503      & 0.444      \\
                             \hline
\multirow{3}{*}{GauS-SLAM}   & PSNR$\uparrow$    & 9.61       & 21.30      & 20.15      & 17.02      \\
                             & SSIM$\uparrow$    & 0.209      & 0.677      & 0.605      & 0.497      \\
                             & LPIPS$\downarrow$ & 0.600      & 0.395      & 0.290      & 0.428      \\
                             \hline
\multirow{3}{*}{GS-ICP SLAM} & PSNR$\uparrow$    & 18.63      & 28.23      & 21.25      & 22.70      \\
                             & SSIM$\uparrow$    & 0.754      & 0.909      & 0.731      & 0.798      \\
                             & LPIPS$\downarrow$ & 0.392      & \cs{0.171} & 0.253      & 0.272      \\
                             \hline
\multirow{3}{*}{GPS-SLAM}    & PSNR$\uparrow$    & \cs{25.87} & \cs{29.34} & \cs{25.14} & \cs{26.79} \\
                             & SSIM$\uparrow$    & \cs{0.875} & \cs{0.929} & \cs{0.811} & \cs{0.871} \\
                             & LPIPS$\downarrow$ & \cs{0.253} & \cf{0.157} & \cf{0.215} & \cs{0.208} \\
                             \hline
\multirow{3}{*}{Ours}        & PSNR$\uparrow$    & \cf{30.64} & \cf{33.12} & \cf{27.11} & \cf{30.29} \\
                             & SSIM$\uparrow$    & \cf{0.939} & \cf{0.931} & \cf{0.856} & \cf{0.909} \\
                             & LPIPS$\downarrow$ & \cf{0.160} & 0.183      & \cs{0.231} & \cf{0.191} \\
\bottomrule
\end{tabular}

\label{table:rq_rird_syn}
\end{table}

\subsection{Per-scene Rendering Quality}
We report detailed per-scene rendering quality results for all evaluated datasets. Tables \ref{table:rq_scannetpp}, \ref{table:rq_rird_syn}, \ref{table:rq_rird_real}, and \ref{table:rq_replica} provide the PSNR, SSIM and LPIPS scores on ScanNet++, RIRD-Syn, RIRD-Real, and Replica, respectively. These results complement the averaged rendering quality reported in the main paper and show the performance of each method on individual scenes.

\begin{table*}[t]
\caption{
Per-scene input-view rendering quality comparison on RIRD-Real. We report PSNR, SSIM, and LPIPS for each real scene and the average across all six scenes.
}
\Description{
A quantitative table comparing input-view rendering quality on the RIRD-Real dataset. The rows list different SLAM methods, and each method is evaluated using PSNR, SSIM, and LPIPS on six real reflective indoor scenes: Living Room 0, Living Room 1, Kitchen, Private Room, Rest Room, and Meeting Room, together with the average performance. The proposed method achieves the best average rendering quality across all reported metrics.
}
\footnotesize
\resizebox{\textwidth}{!}{
\begin{tabular}{c|c|cccccc|c}
\toprule
Method                  & Metrics          & Living Room 0 & Living Room 1 & Kitchen    & Private Room        & Rest Room  & Meeting Room & Average    \\ 
\hline
\multirow{3}{*}{SplaTAM}     & PSNR$\uparrow$    & 24.00         & 26.72         & 24.21      & 24.44      & 22.71      & 14.03        & 22.69      \\
                             & SSIM$\uparrow$    & 0.856         & 0.902         & 0.843      & 0.844      & 0.797      & 0.675        & 0.820      \\
                             & LPIPS$\downarrow$ & 0.352         & 0.281         & 0.336      & 0.277      & 0.378      & 0.507        & 0.355      \\ 
                             \hline
\multirow{3}{*}{RTG-SLAM}    & PSNR$\uparrow$    & 25.42         & 25.39         & 25.16      & 25.34      & \cs{27.25} & \cs{22.42}   & \cs{25.16} \\
                             & SSIM$\uparrow$    & 0.863         & 0.881         & 0.861      & 0.866      & 0.888      & \cs{0.773}   & 0.855      \\
                             & LPIPS$\downarrow$ & 0.365         & 0.345         & 0.371      & 0.302      & 0.317      & \cs{0.455}   & 0.359      \\
                             \hline
\multirow{3}{*}{GauS-SLAM}   & PSNR$\uparrow$    & 25.37         & 26.09         & 26.37      & \cs{27.01} & 25.91      & 17.89        & 24.77      \\
                             & SSIM$\uparrow$    & 0.837         & 0.860         & 0.853      & 0.848      & 0.839      & 0.652        & 0.815      \\
                             & LPIPS$\downarrow$ & 0.306         & 0.306         & 0.305      & 0.264      & 0.342      & 0.456        & 0.330      \\
                             \hline
\multirow{3}{*}{GS-ICP SLAM} & PSNR$\uparrow$    & 26.87         & 27.60         & \cs{27.66} & 26.74      & 27.18      & 13.53        & 24.93      \\
                             & SSIM$\uparrow$    & \cs{0.898}    & \cs{0.916}    & \cs{0.915} & \cs{0.889} & \cs{0.907} & 0.657        & \cs{0.864} \\
                             & LPIPS$\downarrow$ & \cs{0.262}    & \cs{0.237}    & \cs{0.233} & \cs{0.232} & \cf{0.244} & 0.529        & \cs{0.289} \\
                             \hline
\multirow{3}{*}{GPS-SLAM}    & PSNR$\uparrow$    & \cs{27.26}    & \cs{27.70}    & 21.89      & 24.92      & 26.58      & 18.42        & 24.46      \\
                             & SSIM$\uparrow$    & 0.894         & 0.904         & 0.831      & 0.847      & 0.883      & 0.734        & 0.849      \\
                             & LPIPS$\downarrow$ & 0.273         & 0.260         & 0.347      & 0.278      & 0.292      & 0.480        & 0.322      \\
                             \hline
\multirow{3}{*}{Ours}        & PSNR$\uparrow$    & \cf{30.38}    & \cf{31.94}    & \cf{32.56} & \cf{30.56} & \cf{31.15} & \cf{28.35}   & \cf{30.82} \\
                             & SSIM$\uparrow$    & \cf{0.919}    & \cf{0.941}    & \cf{0.937} & \cf{0.919} & \cf{0.923} & \cf{0.886}   & \cf{0.921} \\
                             & LPIPS$\downarrow$ & \cf{0.224}    & \cf{0.190}    & \cf{0.194} & \cf{0.197} & \cs{0.248} & \cf{0.253}   & \cf{0.218} \\
                             \bottomrule
\end{tabular}
}
\label{table:rq_rird_real}
\end{table*}

\begin{table*}[t]
\caption{
Per-scene input-view rendering quality comparison on Replica. We report PSNR, SSIM, and LPIPS for each evaluated scene and the average across all scenes.
}
\Description{
A quantitative table comparing input-view rendering quality on the Replica dataset. The rows list different SLAM methods, and each method is evaluated using PSNR, SSIM, and LPIPS on office and room scenes, together with the average performance. The proposed method remains competitive on this general indoor benchmark.
}
\footnotesize
\resizebox{\textwidth}{!}{
\begin{tabular}{c|c|cccccccc|c}
\toprule
Method                       & Metric & Office 0                     & Office 1                         & Office 2                         & Office 3                         & Office 4                     & Room 0                           & Room 1                       & Room 2                           & Average                     \\ \hline
\multirow{3}{*}{SplaTAM}     & PSNR$\uparrow$  & 38.18                       & 39.26                           & 32.00                           & 30.41                           & 32.41                       & 32.36                           & 33.45                       & 35.01                           & 34.14                       \\
                             & SSIM$\uparrow$  & 0.963                       & 0.958                           & 0.929                           & 0.903                           & 0.916                       & 0.933                           & 0.930                       & 0.952                           & 0.936                       \\
                             & LPIPS$\downarrow$ & 0.118                       & 0.139                           & 0.158                           & 0.166                           & 0.179                       & 0.118                           & 0.133                       & 0.126                           & 0.142                       \\ \hline
\multirow{3}{*}{RTG-SLAM}    & PSNR$\uparrow$  & 37.78                       & 38.41                           & 32.23                           & 31.86                           & 34.98                       & 30.41                           & 32.82                       & 33.82                           & 34.04                       \\
                             & SSIM$\uparrow$  & 0.952                       & 0.950                           & 0.918                           & 0.911                           & 0.935                       & 0.882                           & 0.917                       & 0.929                           & 0.924                       \\
                             & LPIPS$\downarrow$ & 0.141                       & 0.182                           & 0.212                           & 0.197                           & 0.173                       & 0.196                           & 0.179                       & 0.183                           & 0.183                       \\ \hline
\multirow{3}{*}{GauS-SLAM}   & PSNR$\uparrow$  & \cf{42.40} & 39.26                           & \cf{36.18}     & 34.73                           & \cf{40.32} & \cf{39.05}     & \cf{39.55} & \cf{39.33}     & \cf{38.85} \\
                             & SSIM$\uparrow$  & \cf{0.980} & 0.958                           & \cf{0.991}     & \cf{0.969}     & \cf{0.976} & \cf{0.972}     & \cf{0.975} & \cf{0.974}     & \cf{0.974} \\
                             & LPIPS$\downarrow$ & \cf{0.047} & 0.139                           & \cf{0.069}     & \cf{0.059}     & \cf{0.063} & \cf{0.056}     & \cf{0.056} & \cf{0.058}     & \cf{0.068} \\ \hline
\multirow{3}{*}{GS-ICP SLAM} & PSNR$\uparrow$  & 37.79                       & 37.65                           & 29.73                           & 30.57                           & 34.28                       & 28.16                           & 32.88                       & 32.96                           & 33.00                       \\
                             & SSIM$\uparrow$  & 0.964                       & 0.954                           & 0.924                           & 0.932                           & 0.946                       & 0.823                           & 0.930                       & 0.932                           & 0.926                       \\
                             & LPIPS$\downarrow$ & 0.076                       & \cf{0.123}     & 0.132                           & 0.123                           & 0.109                       & 0.128                           & 0.116                       & 0.124                           & 0.116                       \\ \hline
\multirow{3}{*}{GPS-SLAM}    & PSNR$\uparrow$  & 40.91                  & \cs{41.10} & \cs{35.10} & \cs{35.21} & 37.46                  & \cs{34.87} & 36.66                  & \cs{36.67} & 37.25                   \\
                             & SSIM$\uparrow$  & 0.970                    & \cs{0.967}  & 0.947                       & 0.947                      & 0.957                    & 0.947                        & 0.954                    & 0.954                        & 0.955                    \\
                             & LPIPS$\downarrow$ & 0.076                    & 0.132                        & 0.131                        & 0.109                        & 0.103                    & 0.095                        & 0.102                    & 0.114                        & 0.108                    \\ \hline
\multirow{3}{*}{Ours}        & PSNR$\uparrow$  & \cs{41.30} & \cf{41.29}     & 34.79                           & \cf{35.33}     & \cs{37.49} & 34.86                           & \cs{36.67} & 36.59                           & \cs{37.29} \\
                             & SSIM$\uparrow$  & \cs{0.976} & \cf{0.971}     & \cs{0.953}     & \cs{0.954}     & \cs{0.961} & \cs{0.954}     & \cs{0.959} & \cs{0.959}     & \cs{0.961} \\
                             & LPIPS$\downarrow$ & \cs{0.070} & \cs{0.125}     & \cs{0.125}     & \cs{0.103}     & \cs{0.096} & \cs{0.089}     & \cs{0.096} & \cs{0.110}     & \cs{0.102} \\ 
                             \bottomrule
\end{tabular}
}
\label{table:rq_replica}
\end{table*}

\subsection{Tracking Error Visualization on RIRD-Syn}

We further visualize the reconstructed meshes and tracking-error-colored trajectories on RIRD-Syn in Fig.~\ref{fig:tracking_error_vis}. GPS-SLAM~\cite{gps-slam} performs ICP against the TSDF volume, which makes its tracking not affected by reflection artifacts. GS-ICP SLAM~\cite{gs-icp} aligns incoming frames with an optimized Gaussian map. Since reflection effects can disturb the optimization of the Gaussian map, its ICP tracking becomes less reliable in reflective regions. GauS-SLAM~\cite{gaus-slam} uses both color and depth cues for pose estimation and is more sensitive to reflection-induced photometric inconsistency. In contrast, by explicitly modeling reflections, our method achieves more accurate tracking and produces higher-quality mesh reconstruction.


\bibliographystyle{ACM-Reference-Format}
\bibliography{texts/bib}